\documentclass[11pt]{article}

\usepackage[final]{acl}

\usepackage{times}
\usepackage{latexsym}
\usepackage{enumitem}
\usepackage[T1]{fontenc}

\usepackage[utf8]{inputenc}

\usepackage{microtype}

\usepackage{inconsolata}

\usepackage{graphicx}
\usepackage[most]{tcolorbox}
\usepackage{booktabs}
\usepackage{siunitx}
\usepackage{tabularx}

\definecolor{human_ts2v_color}{HTML}{CAD0D7}
\definecolor{human_t2v_color}{HTML}{246B97}
\definecolor{human_ours_color}{HTML}{DE7F29}

\newtoggle{color-macro}
\settoggle{color-macro}{false} %

\iftoggle{color-macro}{
\colorlet{MacroColor}{ETHPetrol}
}{
\colorlet{MacroColor}{black}
}

\usepackage{xurl}
\usepackage{geometry}
\usepackage{booktabs}  
\usepackage[table]{xcolor} 

\definecolor{lightgrey}{gray}{0.92}
\usepackage[normalem]{ulem}
\usepackage{tcolorbox}
\usepackage{tikz}
\usepackage{colortbl}
\usepackage{pgfplots}
\pgfplotsset{compat=1.18}
\usepgfplotslibrary{statistics} %
\usepackage{pgfplotstable}
\usepackage{amsmath, amssymb, amsfonts}
\usepackage{scalefnt}
\usepackage{caption}
\usepackage[font=small,labelfont=bf]{subcaption}
\usepackage{anyfontsize}
\usepackage{multirow}
\usepgfplotslibrary{fillbetween}
\usepackage{changepage}
\usepackage{multicol}
\usepackage{setspace}
\usepackage[version=4]{mhchem}
\usepackage[colorinlistoftodos]{todonotes}
\usepackage{algorithm}
\usepackage{algorithmic}
\usepackage{makecell}
\usepackage{multirow}
\usepackage{collcell}
\usepackage{xfp}    
\usepackage{calc}
\newcommand{\cellbox}[2]{%
  \colorbox{#1}{\makebox[\widthof{000.0}][c]{#2}}%
}
\usepackage{url}
\usepackage{pgfplots}
\usepackage{svg}
\usepgfplotslibrary{fillbetween}
\usetikzlibrary{intersections, pgfplots.statistics}
\pgfplotsset{compat=1.18}
\usepackage{cleveref}
\crefname{section}{\S}{\S\S}
\Crefname{section}{\S}{\S\S}
\crefname{table}{Tab.}{Tabs.}
\crefname{figure}{Fig.}{Figs.}
\crefname{algorithm}{Alg.}{}
\crefname{appendix}{App.}{Apps.}
\crefname{lemma}{Lemma}{}
\Crefname{theorem}{Theorem}{}
\Crefname{assumption}{Assumption}{}
\crefname{proposition}{Proposition}{}
\crefname{hypothesis}{Hypothesis}{}
\crefname{deduction}{Deduction}{}
\crefname{intuition}{\textbf{Intuition}}{\textbf{Intuitions}}
\crefname{observation}{\textbf{Observation}}{\textbf{Observations}}
\crefname{finding}{\textbf{Finding}}{\textbf{Findings}}
\crefname{cor}{Corollary}{}
\crefname{align}{}{}
\crefname{equation}{}{}

\title{\framework: Knowledge-Centric Visual Summarization for Video Lectures}

\author{Yi Xu \\
  City University of Hong Kong \\
  \texttt{yixu96-c@my.cityu.edu.hk} \\\And
  Yifan Hou \\
  ETH Zürich\\
  \texttt{yifan.hou.z@gmail.com} \\\And
  Xiaoyu Zhang\thanks{Corresponding Author} \\
  City University of Hong Kong\\
  \texttt{xiaoyu.zhang@cityu.edu.hk} \\
  }

\newcommand{\framework}{\text{KnowVis}}

\newcommand{\colorImp}[1]{{\color[HTML]{e0791e}#1}}
\newcommand{\colorChg}[1]{{\color[HTML]{755290}#1}}

\begin{document}

\maketitle

\begin{abstract}

Video lectures are valuable educational resources, but their dense and lengthy format often overwhelms novice learners.
This difficulty stems from a fundamental pedagogical mismatch: videos deliver transient information linearly, while human learning requires constructing interconnected cognitive networks, which could induce extreme cognitive overload for novice learners lacking prior domain knowledge.
Existing video summarization methods fail to address this mismatch, as they primarily produce text-heavy, linear condensations that still demand high cognitive effort.
To bridge this gap, we propose $\framework$, a framework that transforms linear video lectures into pedagogically grounded visual narratives.
$\framework$ first extracts a detailed concept map from multimodal video content to identify important and challenging threshold concepts, then constructs structured knowledge units, and finally synthesizes engaging visual summaries. 
Alongside the framework, we introduce a curated dataset of 125 educational videos across 10 academic disciplines, paired with 1,079 generated visual summaries. 
Extensive automated evaluations and a human study demonstrate that, compared to state-of-the-art baselines, $\framework$ generates more accurate and clear visuals that successfully reduce cognitive load and significantly improve student learning effectiveness and knowledge retention.\footnote{Our code and dataset are available at: https://github.com/yixu-cityu/KnowVis. }

\end{abstract}

\begin{figure*}[!t]
  \centering
  \includegraphics[width=\textwidth]{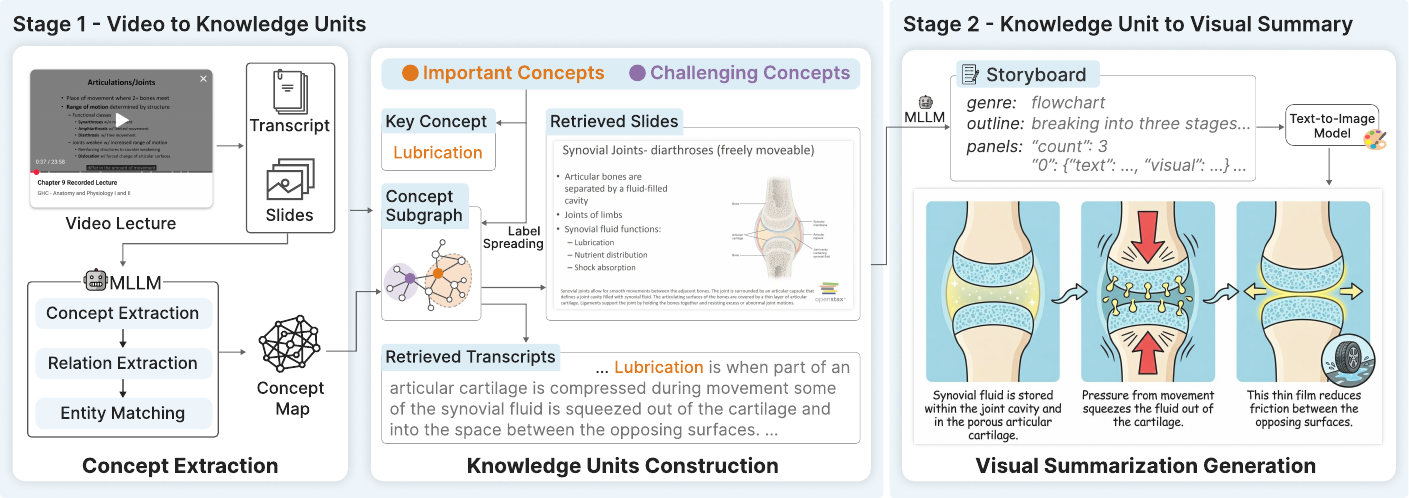}
  \hfill%
  \caption{\textbf{Overview of the \framework~framework.} Stage 1: It extracts a detailed \textit{concept map} from the \textit{video content}, identifies \colorImp{important} and \colorChg{challenging} concepts, and extracts concept subgraphs to construct structured \textit{knowledge units}. Stage 2: It generates storyboards with narratives and visual descriptions to synthesize engaging \textit{visual summaries}.}
  \label{fig:framework}
\end{figure*}

\section{Introduction}
Video lectures have long proven to be a valuable and scalable medium that democratizes access to knowledge from professional instructors across the globe~\citep{ally2004foundations, breslow2013studying, SeatonOnline}.
However, sustaining active engagement during dense, lengthy video content remains a significant challenge~\citep{Ally+2008+15+44}.
For novice learners unfamiliar with a topic, passively watching a continuous stream of information often leads to cognitive fatigue~\citep{DBLP:journals/ce/ChenW15}.
Without clear visual scaffolding, students struggle to extract core concepts, perceive their interrelationships, and maintain focus, which ultimately degrades the overall learning experience~\citep{gutierrez2024video, schacter2015enhancing}.

From a pedagogical perspective, this difficulty stems from a fundamental mismatch between the linear presentation of video media and the non-linear mechanics of human cognitive processing.
According to Cognitive Load Theory~\citep{sweller2020cognitive} and the Cognitive Theory of Multimedia Learning~\citep{mayer1998cognitive,MayerMultimediaLearning}, human learning relies on schema construction: the process of connecting new concepts to existing knowledge networks in a recursive, interconnected manner.
Video lectures, however, deliver information sequentially and transiently.
This ``transient information effect'' forces learners to simultaneously hold earlier concepts in their working memory while processing incoming information, imposing a high \textit{extraneous} cognitive load~\citep{WONG2012449}.
Furthermore, for novice learners, a lack of prior knowledge inherently imposes a high \textit{intrinsic} cognitive load when they attempt to grasp complex concepts~\citep{kalyuga2005prior}.
Consequently, this confluence of high intrinsic and extraneous load overwhelms working memory, actively hindering the learner's ability to meaningfully integrate new knowledge~\citep{swellerCLT}.

Existing computational efforts to mitigate these learning barriers generally fall into three categories, none of which fully resolve this pedagogical mismatch.
First, textual video summarization methods condense lengthy videos into shorter abstracts~\citep{gonzalez-etal-2023-automatically,liu-etal-2025-talk}. However, they still output linear, text-heavy representations that require significant reading effort and fail to visually illustrate the underlying conceptual structures.
Second, video retrieval techniques allow learners to search for and navigate to specific timestamps~\citep{DBLP:journals/tvcg/SchwabSTFHHSKKP17,DBLP:journals/tvcg/ZhouYCWWWCW25,DBLP:conf/chi/LiuKW18}. However, these tools merely aid navigation without synthesizing content to make the underlying knowledge more cognitively accessible.
Finally, while recent advancements in generative AI have enabled domain-specific visual generation, such as transforming math problems into diagrams~\citep{DBLP:conf/acl/WangRWS25} or generating medical flashcards~\citep{DBLP:conf/emnlp/WuGGHLD25}, these approaches rely on heavily constrained, pre-defined templates that cannot generalize to the diverse, multimodal content in open-domain educational lectures.

To bridge the gap between linear video presentation and non-linear cognitive learning, we propose \framework, a framework that automatically generates domain-independent, pedagogically grounded visual summaries from video lectures.
Specifically, the \framework\ pipeline first processes multimodal video inputs (transcripts and slide frames) to extract a comprehensive, non-linear concept map.
Guided by Threshold Concepts theory~\citep{Meek2023}, \framework\ identifies key concepts that are structurally important or cognitively challenging for learners.
The framework then clusters closely related concepts into knowledge units and finally synthesizes narrative visual summaries to explain them.
To support this, we introduce a rigorously curated multimodal dataset comprising 125 video lectures across 10 diverse academic disciplines, paired with 1,079 \framework-generated visual summaries, concept graphs, and source transcripts.

We comprehensively evaluate \framework\ using both automated and human-centric metrics.
Using a Large Language Model (LLM) as an automated judge~\citep{DBLP:conf/nips/ZhengC00WZL0LXZ23}, we demonstrate that \framework\ outperforms state-of-the-art multimodal baselines in accuracy and clarity while optimally balancing information density and significantly minimizing mental effort.
Furthermore, a human study with university students reveals that the visual summaries generated by \framework\ lead to substantially higher learning effectiveness and knowledge retention compared to baseline methods.
Ultimately, this work demonstrates that intelligently transforming linear multimedia into structured visual narratives can significantly reduce cognitive barriers, paving the way for more accessible and engaging self-directed learning experiences.

\section{\framework\ Framework}
In this section, we formulate the task of generating visual summaries from video lectures and introduce our proposed \framework\ framework.

\subsection{Problem Formulation}
The primary objective of \framework\ is to generate visual summaries that facilitate learners' comprehension of complex key concepts within a video lecture.
We formulate this task as a two-stage multimodal summarization problem tailored for educational contexts: $\mathcal{V} \rightarrow \mathcal{K} \rightarrow \mathcal{I}$.

\paragraph{Stage 1: Video Lecture to Knowledge Units ($\mathcal{V} \rightarrow \mathcal{K}$).} 
Given an input video lecture $\mathcal{V}$, this stage constructs a set of discrete knowledge units $\mathcal{K}$ that encapsulate the essential multimodal information surrounding key concepts. 
We first extract a comprehensive concept map $\mathcal{G}$ from $\mathcal{V}$ to capture the lecture's global knowledge structure.
Within this structure, we identify key concepts by isolating structurally important concepts (\colorImp{$\mathcal{C}_{imp}$}) and cognitively challenging concepts (\colorChg{$\mathcal{C}_{chg}$}).
We then construct a set of knowledge units $\mathcal{K}$, modeling each unit $K \in \mathcal{K}$ as the following tuple:
$$K = (c, G, T, S)$$
where $c \in (\colorImp{\mathcal{C}_{imp}} \cup \colorChg{\mathcal{C}_{chg}})$ represents the foundational key concept (e.g., ``Lubrication''). $G \subseteq \mathcal{G}$ is a localized subgraph encoding the concept $c$ and its structurally cohesive relationships within $\mathcal{G}$. $T$ and $S$ denote the specific transcript segments and slide frames, respectively, temporally retrieved from $\mathcal{V}$ to semantically align with $c$.

\paragraph{Stage 2: Knowledge Unit to Visual Summary ($\mathcal{K} \rightarrow \mathcal{I}$).} 
Given a structured knowledge unit $K \in \mathcal{K}$, this stage synthesizes the visual summary $I \in \mathcal{I}$. 
Instead of directly prompting an image generator with raw text, we translate the unit $K$ into a narrative visualization storyboard $B$. This storyboard maps abstract relationships to concrete visual metaphors and spatial layouts, which subsequently guides a text-to-image generative model $g$ to produce the final summary $I = g(B)$.

\subsection{Framework Overview}

As illustrated in Figure~\ref{fig:framework}, \framework\ consists of three core modules. 
First, the \textit{Concept Extraction module} (stage 1) processes multimodal video inputs to extract a fine-grained concept map~$\mathcal{G}$. 
Next, the \textit{Knowledge Units Construction module} (stage 1) evaluates this map to identify important and challenging concepts $(\colorImp{\mathcal{C}_{imp}} \cup \colorChg{\mathcal{C}_{chg}})$ and constructs localized knowledge units $\mathcal{K}$ by aggregating relevant subgraphs, transcripts, and slide frames. 
Finally, the \textit{Visual Summarization Generation module} (stage 2) transforms these structured units into narrative visual summaries $\mathcal{I}$.

\subsection{Concept Extraction}
To build a comprehensive concept map $\mathcal{G}$, we first transcribe the lecture audio and employ a scene transition detection algorithm to isolate unique slide frames. 
To prevent long-context degradation, we segment video transcripts into discrete chunks aligned with these slide transitions. 
For each chunk, we employ a Multimodal Large Language Model (MLLM), i.e., Gemini-3-Flash~\citep{google2025gemini3flash}, taking both the transcript text and the corresponding slide frames as input to extract localized concepts and their relations.

A persistent challenge in automated extraction is entity duplication, where the MLLM extracts semantically identical concepts using different phrasing. 
Inspired by~\citep{DBLP:conf/coling/WangCLCHSWZ25}, we instruct the MLLM to perform entity resolution, aligning identical concepts both within and across temporal chunks. 
Furthermore, to mitigate MLLM hallucination, we apply a strict post-processing filter that prunes any concepts or relations not explicitly grounded in the source video lecture. 
Finally, isolated concept clusters lacking connections to the primary topic graph are removed to maintain narrative conciseness. Detailed extraction prompts are provided in Appendix~\ref{app:sec:concept_map_extraction}.

\begin{figure*}[!t]
  \centering
  \includegraphics[width=\textwidth]{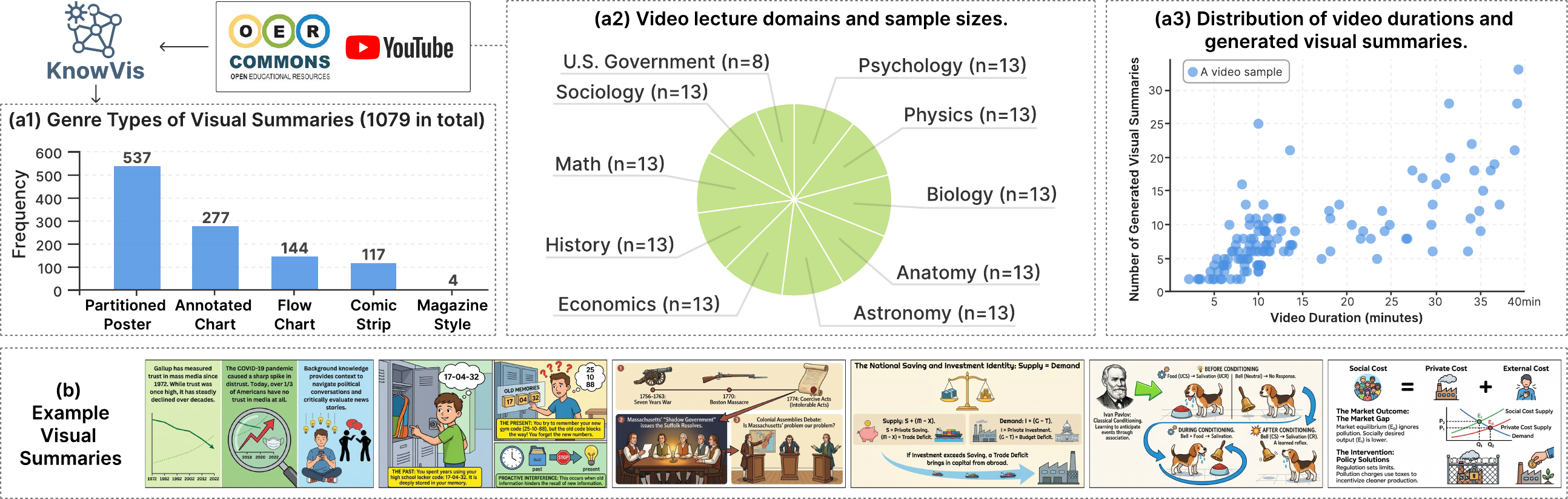}
  \hfill%
  \caption{
  \textbf{Statistics and samples of our Dataset. }
  (a1–a3) Statistics of 1,079 visual summaries generated from 125 video lectures.
  (b) Samples illustrating diverse knowledge concepts (from left to right: ``Trust in media'', ``Proactive interference'', ``Coercive acts'', ``Investment identity'', ``Ivan Pavlov'', ``Social cost'') with pedagogically grounded visual narratives.
  }
  \label{fig:dataset_stats}
\end{figure*}

\subsection{Knowledge Units Construction}
Building upon $\mathcal{G}$, this module identifies critical pedagogical anchors and wraps them into cohesive knowledge units $\mathcal{K}$. Guided by Multimedia Learning Theory~\citep{MayerMultimediaLearning} and the Threshold Concepts framework~\citep{Meek2023}, we categorize key concepts into two dimensions:

\begin{itemize}[leftmargin=*, itemsep=-.2em, topsep=1pt, partopsep=0pt]
    \item \colorImp{Important Concepts.}
    Drawing from signaling principles~\citep{MayerMultimediaLearning}, we identify important concepts based on three criteria: 1) \textit{Signaling}: presence of explicit visual/verbal emphasis; 2) \textit{Temporal Allocation}: the duration of the instructor's focus; and 3) \textit{Structural Connectivity}: the concept's centrality within the global map $\mathcal{G}$. 
    \item \colorChg{Challenging Concepts.}
    Informed by Threshold Concepts Theory~\citep{Meek2023}, we define challenging concepts as those representing ``troublesome knowledge''. These are characterized by counterintuitive logic, highly technical jargon, or multi-step tacit reasoning that frequently disrupts novice comprehension.
\end{itemize}
We input the full transcripts, slides, and $\mathcal{G}$ into the MLLM, prompting it to score each concept across these dimensions. We select the top 10\% of scoring nodes to form our final sets of important ($\mathcal{C}_{imp}$) and challenging ($\mathcal{C}_{chg}$) concepts (see Appendix~\ref{app:sec:img_chg_idenfitication} for more implementation details).

\paragraph{Subgraphs and Information Retrieval.} 
To construct the final knowledge unit $K$, we use each selected key concept $c \in (\colorImp{\mathcal{C}_{imp}} \cup \colorChg{\mathcal{C}_{chg}})$ as a seed node. 
We apply a label-spreading mechanism over $\mathcal{G}$ to cluster closely related neighboring concepts into a localized subgraph $G$. 
Because knowledge is inherently interconnected, concepts may belong to multiple subgraphs if their association probabilities to multiple seeds exceed a threshold of $0.8$ (hyperparameter details in Appendix~\ref{appendix:ku_subgraph}). 
Finally, we input the full transcripts, slides and $G$ into the MLLM to retrieve semantically aligned transcript segments $T$ and slide frames $S$, ensuring contextual completeness for the knowledge unit.

\subsection{Visual Summarization Generation}

Guided by the principles of narrative visualization~\citep{5613452}, this module transforms the abstract data within $K$ into an intuitive visual summary $I$. 
First, we prompt Gemini-3-Flash to analyze content in $K$ to select the most pedagogically suitable visualization genre (e.g., a Flow Chart for procedural knowledge, an Annotated Chart for data, or a Comic Strip for abstract metaphors). 
The MLLM then generates a structured storyboard $B$, mapping the concepts in $G$ to specific visual layouts and descriptive captions.

This storyboard $B$ is passed to Gemini-3.1-Flash-Image~\citep{gemini3_1flashImage} to synthesize the raw image. 
Because generative vision models occasionally produce typographical errors or hallucinated artifacts, we introduce an automated verification step. 
The generated image $I$ is fed back into Gemini-3-Flash alongside the original unit $K$; if semantic or visual inconsistencies are detected, the model refines the storyboard prompt $B$ and triggers a second generation pass to ensure high-fidelity educational accuracy (see Appendix~\ref{appendix:visual_gen}).

\section{Dataset}
\label{sec:dataset}
To validate our framework, we required a domain-diverse, multimodal dataset. 
We curated a comprehensive collection of video lectures sourced from open-access platforms, primarily OER Commons\footnote{\url{https://oercommons.org/}} and YouTube\footnote{\url{https://www.youtube.com/}}. 
To ensure content validity and pedagogical quality, we selected video lectures aligned with the curriculum of OpenStax\footnote{\url{https://openstax.org/}} , a widely recognized open educational resource. 
To support future research, the collected videos and resulting data are openly licensed (see Appendix~\ref{app:sec:data_license_detail}).

As illustrated in Figure~\ref{fig:dataset_stats}, the final dataset comprises 125 video lectures spanning 10 distinct academic disciplines.
From these multimodal inputs, we utilized \framework\ to generate 1,079 pedagogically structured visual summaries. 
Each summary is paired with its source transcripts, slide frames, and concept graphs, providing a rich multimodal resource for future educational generation research.

\section{Automatic Evaluation}
\label{sec:automatic_eval}
Having constructed the dataset, we conduct a large-scale automated evaluation to quantitatively assess the pedagogical quality of the generated visual summaries. 
A case study showing qualitative comparisons is provided in Section~\ref{sec:case_study}.

\subsection{Experimental Setup}
\paragraph{Metrics.}
To quantitatively evaluate the generated visual summaries at scale, we adopt an LLM-as-a-judge approach, leveraging GPT-5.4~\citep{gpt5} as the automated evaluator (See detailed cross-model evaluation in Appendix~\ref{app:claude_results}).
Each visual summary is scored on a discrete scale from 1 to 5 across four pedagogical metrics.
For the first two metrics, a higher score is strictly better:
\begin{itemize}[leftmargin=*, itemsep=-.2em, topsep=1pt, partopsep=1pt]
    \item \textbf{Accuracy:} Measures how faithfully the visual summary aligns with the factual information presented in the source video lecture.
    \item \textbf{Clarity:} Assesses how easily a novice student can interpret the visual summary without confusion or ambiguity~\citep{DBLP:conf/acl/WangRWS25}.
\end{itemize}
For the latter two metrics, the scoring follows a bell-curve preference where 3 is the balance. Scores of 1 (sparse/simple) or 5 (overwhelming/complex) indicate suboptimal pedagogical design:
\begin{itemize}[leftmargin=*, itemsep=-.2em, topsep=1pt, partopsep=1pt]
    \item \textbf{Information Density:} Measures the volume of information presented. Following Multimedia Learning Theory~\citep{MayerMultimediaLearning}, a visual summary should convey critical information (avoiding a score of 1) while omitting redundant details that cause clutter (avoiding a score of 5).
    \item \textbf{Mental Effort:} Assesses the cognitive load required to process the visual. According to Cognitive Load Theory~\citep{swellerCLT}, the summary should foster active processing (avoiding 1) while minimizing extraneous cognitive load (avoiding 5).
\end{itemize}
The detailed prompt used for this LLM-based evaluation is provided in Appendix~\ref{appendix:llm_judge_prompt}.

\paragraph{Baselines.}
To evaluate the efficacy of \framework, we compare its performance against four baseline generation pipelines. 
The baseline pipelines represent different combinations of multimodal input and reasoning: \textbf{V2I} generates summaries directly from the raw video lecture file; \textbf{T2I} relies solely on the textual video transcripts; \textbf{TS2I} utilizes both transcripts and extracted slide frames; and \textbf{TS2I + CoT} incorporates an explicit Chain-of-Thought reasoning step prior to generation based on the transcripts and slides.
To ensure a fair comparison, all baselines are strictly prompted to generate visual summaries for the exact same key concepts identified by our framework, which are clearer and more accurate than those extracted by baseline methods (see Appendix~\ref{app:concept_eval}).

\paragraph{Model Constraints.} 
Because \framework\ utilizes Gemini-3.1-Flash-Image, we comprehensively evaluate this model across all four of the aforementioned pipelines. 
To establish a broader benchmark, we also compare our results against two other state-of-the-art models: Flux-2-Pro~\cite{flux2pro} and Qwen-Image-2~\cite{zhao2026qwenimage20technicalreport}. 
However, these models introduce specific technical constraints. 
Directly inputting video files (V2I) caused Out-Of-Memory (OOM) errors for 29 videos in our dataset, and both Flux and Qwen impose strict limitations on the number of image inputs they can process. 
Consequently, we evaluate Flux and Qwen exclusively using the text-only T2I pipeline. 
Furthermore, due to rigorous internal safety filters, Qwen-Image-2 refused to generate visuals for 51 sensitive concepts, while Flux-2-Pro failed for 442 concepts (primarily in Anatomy and History domains). 
We account for these failures in the results analysis by comparing methods using metric scores averaged over their successful generations.
Appendix~\ref{app:subset_baseline} further provides intersection-based comparisons between \framework{} and these baselines.

\subsection{Experiment Results}

\definecolor{color_3.45-3.6}{HTML}{b2182b}
\definecolor{color_3.3-3.45}{HTML}{d6604d}
\definecolor{color_3.15-3.3}{HTML}{f4a582}
\definecolor{color_3.0-3.15}{HTML}{fddbc7}
\definecolor{color_2.85-3.0}{HTML}{f7f7f7}
\definecolor{color_2.7-2.85}{HTML}{d1e5f0}
\definecolor{color_2.55-2.7}{HTML}{92c5de}
\definecolor{color_2.4-2.55}{HTML}{4393c3}
\definecolor{color_2.25-2.4}{HTML}{2166ac}

\begin{table*}[!h]
\centering
\small
\caption{\textbf{Quantitative evaluation of automated visual summary generation methods.} We compare direct and Chain-of-Thought (CoT) prompting across Flux, Qwen, and Gemini using various input modalities against our proposed \framework{}. \framework{} consistently outperforms baseline methods, achieving the highest accuracy and clarity while optimally filtering information density and significantly minimizing learner mental effort.}

\label{tab:baseline_results}
\vspace{0.5em}

\begin{tabular}{l S[table-format=1.3] S[table-format=1.3] S[table-format=1.3] S[table-format=1.3]}
\toprule
\midrule
Method & {\textbf{Accuracy $\uparrow$}} & {\textbf{Clarity $\uparrow$}} & {\textbf{Information Density}} & {\textbf{Mental Effort}} \\
\midrule
Flux-T2I           & 3.080 & 3.443 & {\cellbox{color_3.0-3.15!60}{3.047}} & {\cellbox{color_3.0-3.15!60}{3.055}} \\ 

Qwen-T2I           & 3.618 & 4.152 & {\cellbox{color_2.85-3.0!60}{2.994}} & {\cellbox{color_2.7-2.85!60}{2.764}} \\

Gemini-T2I         & 3.716 & 4.144 & {\cellbox{color_3.3-3.45!60}{3.442}} & {\cellbox{color_3.0-3.15!60}{3.014}} \\ 

Gemini-V2I         & 3.564 & 4.102 & {\cellbox{color_3.3-3.45!60}{3.335}} & {\cellbox{color_2.85-3.0!60}{2.981}} \\ 

Gemini-TS2I        & 3.721 & 4.057 & {\cellbox{color_3.45-3.6!60}{3.471}} & {\cellbox{color_3.0-3.15!60}{3.083}} \\ 

Gemini-TS2I (CoT)  & 3.682 & 4.005 & {\cellbox{color_3.45-3.6!60}{3.570}} & {\cellbox{color_3.15-3.3!60}{3.201}} \\ 
\midrule
\framework{}              & {\bfseries 3.723} & {\bfseries 4.470} & {\cellbox{color_2.7-2.85!60}{2.816}} & {\cellbox{color_2.25-2.4!60}{2.366}} \\ 
\midrule
\bottomrule
\end{tabular}

\end{table*}

\paragraph{Accuracy and Clarity.} 
We first evaluate the fundamental quality of the generated summaries. 
As shown in Table~\ref{tab:baseline_results}, \framework\ achieves the highest accuracy score ($3.723$), narrowly outperforming the top Gemini baselines. 
More notably, \framework\ yields a substantial improvement in clarity ($4.470$), leading all baseline methods by a wide margin. 
This significant gain highlights a core flaw in standard pipelines: baseline methods tend to dump all retrieved text and slide details into the prompt without prioritization, resulting in unfocused and chaotic images. 
By structuring knowledge units into a logical narrative flow before image synthesis, \framework\ aims to ensure that the core educational message remains focused and unambiguous.

\paragraph{Information Density.} 
Next, we examine the volume of information presented, where a score of 3.0 represents the middle pedagogical density. 
While Qwen-T2I ($2.994$) and Flux-T2I ($3.047$) hover closest to this median, \framework\ scores slightly lower at 2.816. 
Rather than a deficiency, this conciseness is a deliberate and highly beneficial feature of our design. 
Because \framework\ distills lengthy transcripts into abstract concept graphs \textit{before} generating the narrative, it inherently strips away verbal fluff. 
By prioritizing a coherent narrative structure over a high density of fine-grained details, the framework prevents the visual clutter that typically overwhelms novice learners.

\paragraph{Mental Effort.}
Finally, we assess the cognitive load imposed on learners. 
The visual summaries generated by \framework\ require significantly lower mental effort ($2.366$) to process than all other evaluated methods. 
Conversely, applying a CoT reasoning step to the TS2I pipeline generates the most mentally demanding summaries ($3.201$), as the model attempts to visualize overly complex logical chains. 
The success of \framework\ lies in its synthesis of the previous metrics: high accuracy and clarity, combined with concisely filtered information density, naturally minimize cognitive fatigue. 
By wrapping facts in a structured narrative design, \framework\ actively reduces extraneous cognitive load, allowing learners to comprehend complex concepts with far less mental strain.

\section{Ablation Study}
\label{sec:ablation}
To understand the function of each module in our pipeline, we conducted an ablation study. 
By systematically removing key modules, we can isolate their specific impact on pedagogical quality.

\subsection{Experiment Setup}
To understand the contribution of each module in our pipeline, we conducted an ablation study using Gemini-3-Flash and Gemini-3.1-Flash-Image across three degraded settings:
\begin{enumerate}[leftmargin=*, itemsep=-.2em, topsep=0pt, partopsep=1pt]
    \item \textbf{w/o Concept Extraction:} We remove the structured map extraction, directly prompting the MLLM to extract text-based concepts from the transcripts and slides.
    \item \textbf{w/o Knowledge Units Construction:} We skip the important/challenging filtering and label-spreading. The MLLM directly extracts subgraphs, retrieves text and slides to build units.
    \item \textbf{w/o Visual Summarization Generation:} We bypass the narrative storyboard design, directly prompting the model with knowledge units.
\end{enumerate}

\subsection{Experiment Results}
This section examines the role of each module by comparing metric scores averaged over generated visual summaries in each ablation setting (Table~\ref{tab:ablation_results}). A supplementary comparison based on video-level averages is provided in Appendix~\ref{app:ablation_video_level}.

\definecolor{color_3.45-3.6}{HTML}{b2182b}
\definecolor{color_3.3-3.45}{HTML}{d6604d}
\definecolor{color_3.15-3.3}{HTML}{f4a582}
\definecolor{color_3.0-3.15}{HTML}{fddbc7}
\definecolor{color_2.85-3.0}{HTML}{f7f7f7}
\definecolor{color_2.7-2.85}{HTML}{d1e5f0}
\definecolor{color_2.55-2.7}{HTML}{92c5de}
\definecolor{color_2.4-2.55}{HTML}{4393c3}
\definecolor{color_2.25-2.4}{HTML}{2166ac}
\begin{table*}[!h]
\centering
\small
\caption{\textbf{Quantitative ablation study results.} We evaluate the impact of removing key structural modules from the \framework{} pipeline. The results demonstrate that each component plays a necessary role to balance high accuracy and clarity while maintaining optimal pedagogical cognitive load.}
\label{tab:ablation_results}
\vspace{0.5em}
\begin{tabularx}{0.85\textwidth}{>{\raggedright\arraybackslash}X c *{4}{S[table-format=1.3]}} 
\toprule
\midrule
{\makecell[l]{\textbf{Setting}}} & {\makecell[c]{Num.\textsuperscript{a}}} & {\makecell[c]{Accuracy $\uparrow$ }} & {\makecell[c]{Clarity $\uparrow$}} & {\makecell[c]{Information\\Density}} & {\makecell[c]{Mental\\Effort}} \\
\midrule
\framework\                         & 1079 & 3.723 & 4.470 & {\cellbox{color_2.7-2.85!60}{2.816}} & {\cellbox{color_2.25-2.4!60}{2.366}} \\
w/o Concept Extraction              & 724  & 3.738 & 4.354 & {\cellbox{color_2.7-2.85!60}{2.746}} & {\cellbox{color_2.25-2.4!60}{2.365}} \\
w/o Knowledge Units Construction     & 1401 & 3.717 & 4.481 & {\cellbox{color_2.85-3.0!60}{2.857}} & {\cellbox{color_2.25-2.4!60}{2.349}} \\
w/o Visual Summarization Generation  & 1079 & 3.702 & 4.068 & {\cellbox{color_3.3-3.45!60}{3.428}} & {\cellbox{color_3.0-3.15!60}{3.063}} \\
\midrule
\bottomrule
\end{tabularx}

\vspace{0.2em}
{\raggedright \small \textsuperscript{a} {Number of visual summaries generated.\par}}
\end{table*}

\paragraph{The Role of Concept Extraction.}
We first evaluate the necessity of the global concept map. As shown in Table~\ref{tab:ablation_results}, without it, accuracy artificially rises slightly ($3.723 \rightarrow 3.738$), but Clarity drops ($4.470 \rightarrow 4.354$). Furthermore, the pipeline generated only 724 visual summaries compared to the baseline 1,079. When forced to extract concepts directly without a map, the MLLM outputs lengthy, convoluted sentences instead of precise keywords. While this smaller pool of broad content mathematically reduces the chance for generation errors (thereby raising Accuracy), these lengthy text strings fail to act as precise visual anchors. Consequently, the lack of structural precision significantly degrades the Clarity of the final image.

\paragraph{The Role of Knowledge Units Construction.}
Next, we assess the pedagogical filtering module. Skipping construction of key concepts results in massive, highly fragmented visual summaries (n=1,401). 
Because these summaries represent simple, isolated facts rather than structural knowledge, they score highest in Clarity ($4.481$) and lowest in Mental Effort ($2.349$), but suffer a drop in Accuracy ($3.723\rightarrow3.717$). While isolated facts are superficially easier to read, they lack true pedagogical value. 
The Knowledge Unit module is critical because it forces the framework to anchor visual generation around threshold concepts, ensuring the summaries guide learners through actual, structured knowledge rather than disjointed trivia.

\paragraph{The Role of Narrative Visual Generation.} 
Finally, we isolate the impact of narrative design by removing the Visual Summarization module. This removal causes a severe degradation in Clarity ($4.470 \rightarrow 4.068$) while drastically increasing Information Density ($2.816 \rightarrow 3.428$) and Mental Effort ($2.366 \rightarrow 3.063$). This demonstrates that directly feeding structured data to an image generator produces cluttered, overwhelming visuals. The intermediary step of designing a narrative storyboard is strictly essential to translate raw data into an intuitive, cognitively accessible format.

\section{Human Evaluation}
\label{sec:human_eval}
To evaluate the effectiveness of \framework\ in real-world learning environments, we conducted a user study with university students to capture authentic human interpretation of the generated visual summaries.

\subsection{Experiment Setup}
We recruited 10 university students ($P_1$--$P_{10}$) with diverse academic backgrounds, including Computer Science, Bioengineering, Architecture, Linguistics, Law, and Literature. 
The participants were randomly divided into five stable dyadic groups ($G_1$--$G_5$). 
We then randomly selected one video lecture from each of the 10 disciplines in our dataset, partitioned them into five pairs, and randomly assigned one pair to each group.

For each group, the two evaluators independently watched their assigned lectures and evaluated the corresponding visual summaries generated by \framework\ against two baseline configurations: T2I and TS2I using Gemini-3.1-Flash-Image. 
These baselines were selected as they demonstrated the strongest unimodal and multimodal performance, respectively, in our automated evaluation. 
The presentation order of the options was randomized, and evaluators were blind to the underlying methods to eliminate bias.
Evaluators assessed the generation outputs across three core pedagogical dimensions:
\begin{itemize}[leftmargin=*, itemsep=-.2em]
    \item \textbf{Learning Effectiveness:} Measures how efficiently a student can comprehend the underlying target concept through the visual summary.
    \item \textbf{Knowledge Retention:} Assesses how reliably the core educational insights of the concept can be memorized and recalled via the visuals.
    \item \textbf{Knowledge Transfer:} Measures a student's willingness to share the visual summary with their peers as an illustrative learning aid.
\end{itemize}
Participants identified the best and worst visual summaries among the choices for each concept. 
To quantitatively analyze these subjective preferences, we mapped the categorical rankings to numerical scores: 5 for the best summary, 0 for the worst, and 3 for the intermediate runner-up. 
An example of the human evaluation interface is in Appendix~\ref{app:sec:human_eval}.

\begin{figure}[!ht]
	\noindent
	\begin{tikzpicture}
	\begin{axis}[
            trim axis left,
	        ybar, 
	        bar width=0.25cm, 
	        width=7.0cm,    
	        height=5.0cm,     
            ymin=0, ymax=4,
	        enlarge x limits=0.25, 
	        enlarge y limits={upper, value=0.2}, 
	        ylabel={ \footnotesize{Score}},
            y label style={font=\footnotesize},
	        symbolic x coords={LE, KR, KT},
	        xtick={LE, KR, KT},
            xtick align = inside,
	        xticklabels={Learning\\Effectiveness, Knowledge\\Retention, Knowledge\\Transfer},
	        x tick label style={font=\small, align=center}, 
            y tick label style={font=\footnotesize},
	        xlabel near ticks,
	        ylabel near ticks,
	        xmajorgrids=true,
	        ymajorgrids=true,
	        grid style=dashed,
	        legend style={at={(0.5,0.86)}, anchor=south, nodes={scale=0.7, transform shape},draw=none},
	        legend columns=3, 
            legend image code/.code={
                \draw[draw=none] (0cm,-0.1cm) rectangle (0.3cm,0.1cm);
            }
	    ]
            \addplot+[line width=0.4mm, color=human_ts2v_color, 
                error bars/y dir=both, 
                error bars/y explicit, 
                error bars/error bar style={color=gray, line width=0.5pt}, 
                error bars/error mark options={rotate=90, mark size=1.5pt, color=gray} 
            ] coordinates {(LE, 2.0517) +- (0, 0.1908) (KR, 1.8017) +- (0, 0.1937) (KT, 2.1897) +- (0, 0.1966)};
	        \addlegendentry{Gemini TS2I}
            
	        \addplot+[line width=0.4mm, color=human_t2v_color, 
                error bars/y dir=both, 
                error bars/y explicit, 
                error bars/error bar style={color=gray, line width=0.5pt}, 
                error bars/error mark options={rotate=90, mark size=1.5pt, color=gray} 
            ] coordinates {(LE, 2.8966) +- (0, 0.1846) (KR, 2.7759) +- (0, 0.1648) (KT, 3.0086) +- (0, 0.1747)};

	        \addlegendentry{Gemini T2I}
	         \addplot+[line width=0.4mm, color=human_ours_color, 
                error bars/y dir=both, 
                error bars/y explicit, 
                error bars/error bar style={color=gray, line width=0.5pt}, 
                error bars/error mark options={rotate=90, mark size=1.5pt, color=gray} 
            ] coordinates {(LE, 3.0517) +- (0, 0.1860) (KR, 3.4224) +- (0, 0.1841) (KT, 2.8017) +- (0, 0.1945)};
	        \addlegendentry{Ours}
	\end{axis}
	\end{tikzpicture}
	\caption{Average human-evaluated scores for generated visual summaries across Learning Effectiveness, Knowledge Retention, and Knowledge Transfer.
    }
	\label{fig:human_eval_result}
\end{figure}
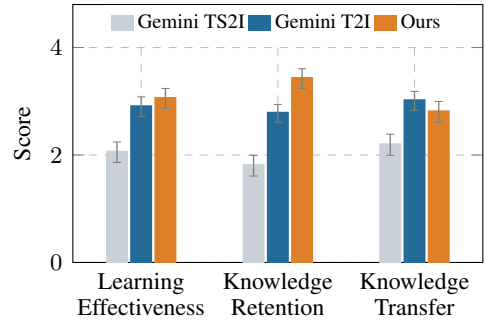

\subsection{Experiment Results}
As illustrated in Figure~\ref{fig:human_eval_result}, \framework\ comfortably outperforms both baseline methods across the primary educational metrics. 
Specifically, for Learning Effectiveness, our framework achieves the highest average score (M=3.052, SE=0.186), followed by the T2I baseline (M=2.897, SE=0.185) and the TS2I configuration (M=2.052, SE=0.191). 
We observe a more pronounced advantage in Knowledge Retention, where \framework\ yields a clear lead (M=3.422, SE=0.184) over T2I (M=2.776, SE=0.165) and TS2I (M=1.802, SE=0.194). 
These findings corroborate our automated evaluation: by anchoring the visual generation around structured knowledge units and an explicit narrative flow, \framework\ provides sticky, low-effort cognitive landmarks that actively help students unpack and retain complex concepts.
For the Knowledge Transfer metric, however, the text-only T2I baseline obtains a slightly higher average preference score (M=3.009, SE=0.175) than our framework (M=2.802, SE=0.195), while TS2I scores lowest (M=2.190, SE=0.197). 
To diagnose this variance, we analyze the inter-rater reliability below.

\begin{table}[!t]
\centering
\caption{Inter-Rater Consistency Evaluated by Weighted Cohen's Kappa Across Participants.}
\label{tab:consistency_check}
\vspace{0.5em}
\small
\setlength{\tabcolsep}{2pt} 
\resizebox{\columnwidth}{!}{
    \begin{tabular}{cc l @{\hspace{0pt}} ccc}
    \toprule
    \midrule
    \textbf{Group} & \textbf{ID} & \textbf{Major} & \makecell[c]{\textbf{Learning}\\\textbf{Effectiveness}\\($\kappa$)} & \makecell[c]{\textbf{Knowledge}\\\textbf{Retention}\\($\kappa$)} & \makecell[c]{\textbf{Knowledge}\\\textbf{Transfer}\\($\kappa$)} \\
    \midrule
    
    \multirow{2}{*}{$G_1$} & $P_1$ & Linguistic & \multirow{2}{*}{0.455} & \multirow{2}{*}{0.932} & \multirow{2}{*}{0.727} \\ 
                        & $P_2$ & Law        &                        &                        &                        \\ 
    \addlinespace[0.5em] %
    
    \multirow{2}{*}{$G_2$} & $P_3$ & Computer Science & \multirow{2}{*}{$-0.096$} & \multirow{2}{*}{$-0.096$} & \multirow{2}{*}{0.077} \\ 
                        & $P_4$ & Linguistic       &                           &                           &                        \\ 
    \addlinespace[0.5em]
    
    \multirow{2}{*}{$G_3$} & $P_5$ & Literature    & \multirow{2}{*}{0.250} & \multirow{2}{*}{0.156} & \multirow{2}{*}{0.156} \\ 
                        & $P_6$ & Public Health &                        &                        &                        \\ 
    \addlinespace[0.5em]
    
    \multirow{2}{*}{$G_4$} & $P_7$ & Bioengineering & \multirow{2}{*}{0.386} & \multirow{2}{*}{0.250} & \multirow{2}{*}{0.250} \\ 
                        & $P_8$ & Architecture           &                        &                        &                        \\ 
    \addlinespace[0.5em]
    
    \multirow{2}{*}{$G_5$} & $P_9$  & Linguistic       & \multirow{2}{*}{0.786} & \multirow{2}{*}{0.464} & \multirow{2}{*}{$-0.286$} \\ 
                        & $P_{10}$ & Computer Science &                        &                        &                           \\ 
    \midrule
    \bottomrule
    \end{tabular}
}

\end{table}

\subsection{Consistency Analysis}
We assessed inter-rater reliability using a linear-weighted Cohen's Kappa ($\kappa$). 
As shown in Table~\ref{tab:consistency_check}, Learning Effectiveness and Knowledge Retention maintain acceptable consistency across most dyads. 
Where divergence occurs, it is primarily driven by disparities in prior knowledge. For instance, in $G_2$ ($\kappa=-0.096$), an evaluator with high domain familiarity ($P_4$) preferred information-dense baselines to capture fine details, whereas a novice co-evaluator ($P_3$) heavily favored \framework's concise narratives to quickly grasp high-level concepts.

Inter-rater alignment drops further for Knowledge Transfer (e.g., $\kappa=-0.286$ in $G_5$), which post-study interviews attribute to competing social curation philosophies rather than pedagogical quality. 
While some students ($P_9$) prefer sharing \framework's lean summaries to spare peers from cognitive fatigue, others ($P_{10}$) share dense baseline visuals purely out of caution that peers might miss secondary details. 
This social dichotomy explains why text-heavy baselines marginally edge out \framework\ in raw transfer scores, despite lagging significantly in standalone educational utility.

\section{Case Study}
\label{sec:case_study}

\begin{figure*}[!h]
  \centering
  \includegraphics[width=\textwidth]{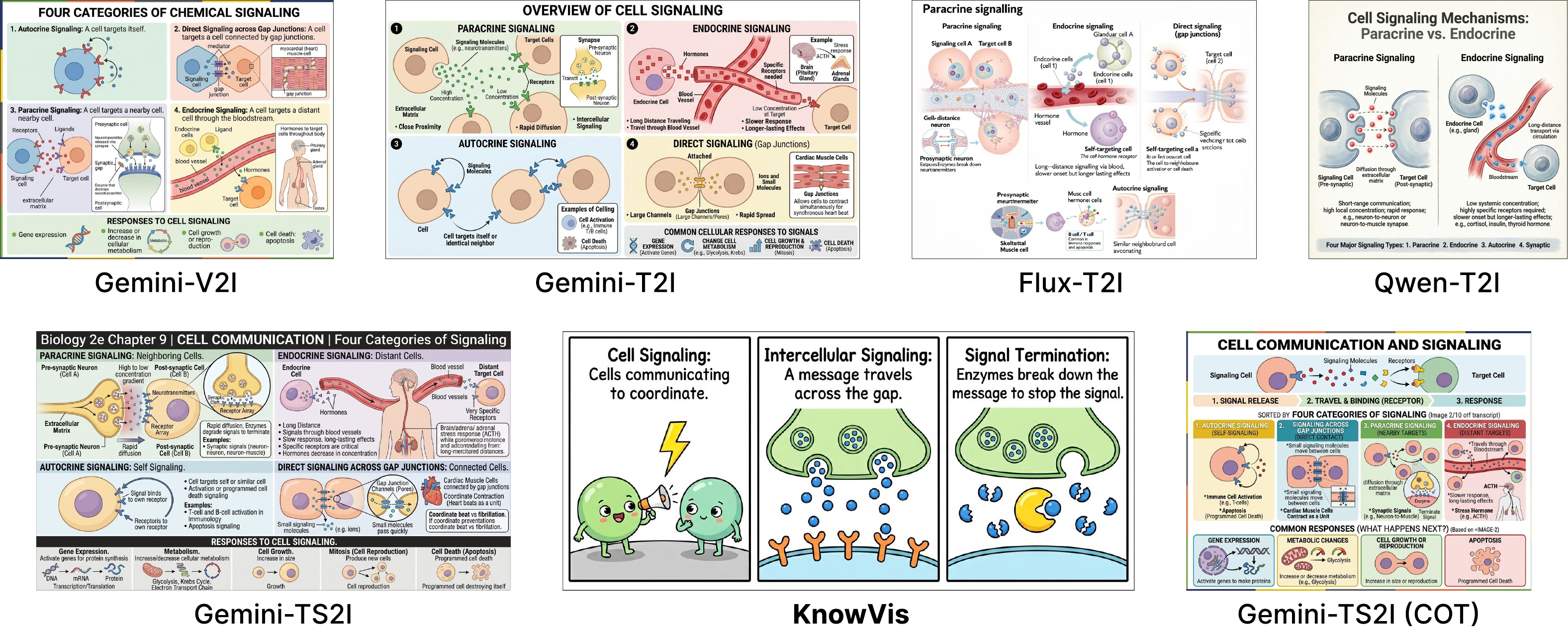}
  \hfill%
  \caption{Qualitative comparison of visual summaries generated by different methods: direct and Chain-of-Thought (CoT) prompting across Flux, Qwen, and Gemini using various input modalities against our proposed \framework{}. KnowVis output optimizes learning effectiveness by delivering a concise, engaging visual narrative that is less cognitively demanding for learners.}
  \label{app:fig:case_study}
\end{figure*}

To qualitatively evaluate the visual summaries, we compare representative examples generated by different methods.
As shown in Figure~\ref{app:fig:case_study},
\framework~generates high-quality, pedagogically meaningful visual summaries that excel across multiple dimensions.
First, compared with the baseline methods, \framework~presents clearer and more intuitive textual and visual elements, enabling learners to identify essential information more efficiently.
Next, it provides an appropriate amount of information without overwhelming details, avoiding the cognitive overload triggered by text-heavy diagrams.
Furthermore, the content is highly accessible and easy to understand for novice learners. 
By integrating relatable real-world metaphors and using clear language, it bridges the gap between complex knowledge and intuitive understanding, maximizing its educational effectiveness for learners with diverse backgrounds.

\section{Related Work}
\label{sec:related_work}

\paragraph{Pedagogical Foundations.}
The theoretical foundation of our work stems from the Cognitive Theory of Multimedia Learning and Cognitive Load Theory~\citep{swellerCLT,sweller2020cognitive,mayer1998cognitive,MayerMultimediaLearning}. 
These theories demonstrate that processing linear instructional formats often overwhelms working memory, and that learning is optimized when verbal and visual representations are actively integrated. 
To systematically identify which concepts require this visual scaffolding, educational researchers rely on the Threshold Concepts framework~\citep{Meek2023} to pinpoint ``troublesome knowledge'' that acts as a bottleneck for novices. 
Our framework directly operationalizes these pedagogical theories to guide automated visual generation.

\paragraph{Computational Approaches.}
Existing computational efforts to mitigate learning barriers in video lectures generally fall into three categories. 
First, textual video summarization condenses multimedia content into text abstracts, either by generating global textual overviews~\citep{gonzalez-etal-2023-automatically,liu-etal-2025-talk,DBLP:conf/ijcnlp/PennecLAMC25}, segment-anchored summaries~\citep{DBLP:conf/cvpr/LeeGC25,DBLP:journals/tmm/LinHCLHHL24}, or synthesizing question-answer pairs that capture important topics~\cite{DBLP:conf/acl/DingZWJY25,DBLP:conf/emnlp/RaySAG25}.
While this reduces overall length, it still outputs linear, text-heavy representations that fail to visually illustrate structural concept relationships. 
Second, video retrieval and indexing techniques enable learners to navigate to specific timestamps~\citep{DBLP:journals/tvcg/SchwabSTFHHSKKP17,DBLP:journals/tvcg/ZhouYCWWWCW25,DBLP:conf/chi/LiuKW18}. 
However, these tools merely aid navigation without actively synthesizing the underlying knowledge. 
Finally, recent generative AI studies have explored domain-specific visual generation, such as math diagrams or medical flashcards~\citep{DBLP:conf/emnlp/WuGGHLD25,DBLP:conf/acl/WangRWS25}. 
While educationally effective, these approaches rely on heavily constrained, pre-defined templates that cannot generalize to the diverse, open-domain multimodal content of general lectures.

\paragraph{Our Position.}
Our proposed \framework\ addresses these limitations by bridging open-domain multimodal processing with pedagogically grounded visual synthesis. 
Unlike textual summarizers or navigational indices, \framework\ actively restructures linear video content into non-linear concept maps to expose underlying relationships. 
Furthermore, unlike domain-restricted visual generators, our framework utilizes MLLMs to dynamically design narrative visual summaries without relying on rigid templates, offering a scalable solution to reduce cognitive load in open-domain learning.

\section{Conclusion}
\label{sec:conclusion}
In this paper, we introduced \framework, a domain-independent framework that transforms lengthy, linear video lectures into pedagogically grounded visual summaries. 
By leveraging MLLMs to extract structural concept maps and identify challenging educational anchors, our pipeline generates intuitive narrative summarizations that explicitly map complex knowledge relationships. 
Extensive automated evaluations and human studies demonstrate that \framework\ significantly outperforms existing baselines, enhancing clarity and accuracy while effectively minimizing extraneous cognitive load. 
Ultimately, this work provides a scalable, multimodal solution to alleviate cognitive fatigue for novice learners, paving the way for more accessible and engaging self-directed education.

\section*{Limitations}
\label{sec:limitations}

While our framework demonstrates strong potential for educational visual generation, we acknowledge several limitations that provide avenues for future research:

\paragraph{Domain Variance (STEM vs. Non-STEM).} 
Although \framework\ is designed to be domain-independent, generating visual metaphors for highly abstract concepts in the Humanities or Social Sciences (e.g., Sociology, History) is inherently more subjective than visualizing concrete physical phenomena in STEM disciplines (e.g., Anatomy, Physics). In our current study, we observe that the visual generation models occasionally struggle to map abstract non-STEM concepts to intuitive graphical layouts. Future work should systematically investigate these cross-domain discrepancies and explore domain-adaptive prompting strategies.

\paragraph{Dependence on Proprietary Models.} 
Our current pipeline relies on closed-source, proprietary models (Gemini-3-Flash and Gemini-3.1-Flash-Image) accessed via APIs. While these models represent the state-of-the-art in multimodal reasoning and generation, their closed nature introduces potential reproducibility challenges, as the models may be subject to silent updates. Furthermore, this limits the community's ability to fine-tune the underlying model weights on specific educational visual datasets. Future iterations should explore deploying open-weight MLLMs to ensure long-term reproducibility and adaptability.

\paragraph{Scale of Human Evaluation.} 
Our user study provided rich qualitative insights regarding cognitive load, prior knowledge impacts, and social curation behaviors. However, the study was conducted with a relatively small sample size (N=10 university students) in a controlled setting. This limits the statistical generalizability of the human evaluation results. A critical next step is to conduct a large-scale, longitudinal deployment of \framework\ in real-world classroom environments to assess its impact on diverse learner populations over extended periods.

\paragraph{Visual and Textual Generative Artifacts.} 
Despite implementing a closed-loop MLLM verification step to refine the storyboards, current text-to-image generative models still occasionally produce artifacts. These include typographic errors (e.g., misspelled text within an annotated chart) and minor spatial hallucinations. Fully eliminating these artifacts remains an open challenge in the broader generative AI field and currently requires human-in-the-loop validation for deployment in high-stakes educational settings.

\paragraph{LLM-as-a-Judge Bias.} 
In our automatic evaluation, we utilized GPT-5.4 to conduct large-scale automated evaluations of our visual summaries. While recent literature validates this approach, LLM judges can exhibit inherent biases, such as preferring specific formatting styles or verbosity levels. Although our human evaluation corroborated the automated trends, the absolute automated scores should be interpreted as strong directional indicators rather than perfect objective truth.

\vspace{2em}
\section*{Acknowledgements}

We thank the participants who volunteered for our user studies and the anonymous reviewers for their feedback and suggestions that helped improve the quality of this paper. This research is sponsored by the CityUHK Start-Up Grant (Project No. 9610708).

\bibliography{main}

@article{ally2004foundations,
  title={Foundations of educational theory for online learning},
  author={Ally, Mohamed},
  journal={Theory and practice of online learning},
  volume={2},
  number={1},
  pages={15--44},
  year={2004}
}

@article{SeatonOnline,
author = {Seaton, Daniel T. and Bergner, Yoav and Chuang, Isaac and Mitros, Piotr and Pritchard, David E.},
title = {Who does what in a massive open online course?},
year = {2014},
issue_date = {April 2014},
publisher = {Association for Computing Machinery},
address = {New York, NY, USA},
volume = {57},
number = {4},
issn = {0001-0782},
url = {https://doi.org/10.1145/2500876},
doi = {10.1145/2500876},
journal = {Commun. ACM},
month = apr,
pages = {58–65},
numpages = {8}
}

@inproceedings{DBLP:conf/acl/DingZWJY25,
  author       = {Wenjian Ding and
                  Yao Zhang and
                  Jun Wang and
                  Adam Jatowt and
                  Zhenglu Yang},
  editor       = {Wanxiang Che and
                  Joyce Nabende and
                  Ekaterina Shutova and
                  Mohammad Taher Pilehvar},
  title        = {Generating Questions, Answers, and Distractors for Videos: Exploring
                  Semantic Uncertainty of Object Motions},
  booktitle    = {Findings of the Association for Computational Linguistics, {ACL} 2025,
                  Vienna, Austria, July 27 - August 1, 2025},
  series       = {Findings of {ACL}},
  pages        = {7207--7220},
  publisher    = {Association for Computational Linguistics},
  year         = {2025},
  url          = {https://aclanthology.org/2025.findings-acl.376/},
  bibsource    = {dblp computer science bibliography, https://dblp.org}
}

@article{DBLP:journals/tmm/LinHCLHHL24,
  author       = {Jingyang Lin and
                  Hang Hua and
                  Ming Chen and
                  Yikang Li and
                  Jenhao Hsiao and
                  Chiuman Ho and
                  Jiebo Luo},
  title        = {VideoXum: Cross-Modal Visual and Textural Summarization of Videos},
  journal      = {{IEEE} Trans. Multim.},
  volume       = {26},
  pages        = {5548--5560},
  year         = {2024},
  url          = {https://doi.org/10.1109/TMM.2023.3335875},
  doi          = {10.1109/TMM.2023.3335875},
  bibsource    = {dblp computer science bibliography, https://dblp.org}
}

@inproceedings{DBLP:conf/ijcnlp/PennecLAMC25,
  author       = {Galann Pennec and
                  Zhengyuan Liu and
                  Nicholas Asher and
                  Philippe Muller and
                  Nancy F. Chen},
  editor       = {Kentaro Inui and
                  Sakriani Sakti and
                  Haofen Wang and
                  Derek F. Wong and
                  Pushpak Bhattacharyya and
                  Biplab Banerjee and
                  Asif Ekbal and
                  Tanmoy Chakraborty and
                  Dhirendra Pratap Singh},
  title        = {Integrating Video and Text: {A} Balanced Approach to Multimodal Summary
                  Generation and Evaluation},
  booktitle    = {Proceedings of the 14th International Joint Conference on Natural
                  Language Processing and the 4th Conference of the Asia-Pacific Chapter
                  of the Association for Computational Linguistics, {IJCNLP-AACL} 2025,
                  Mumbai, India, December 20-24, 2025},
  pages        = {2403--2426},
  publisher    = {The Asian Federation of Natural Language Processing and The Association
                  for Computational Linguistics},
  year         = {2025},
  url          = {https://aclanthology.org/2025.ijcnlp-long.129/},
  bibsource    = {dblp computer science bibliography, https://dblp.org}
}

@ARTICLE{5613452,
  author={Segel, Edward and Heer, Jeffrey},
  journal={IEEE Transactions on Visualization and Computer Graphics}, 
  title={Narrative Visualization: Telling Stories with Data}, 
  year={2010},
  volume={16},
  number={6},
  pages={1139-1148},
  doi={10.1109/TVCG.2010.179}}

@inproceedings{DBLP:conf/emnlp/RaySAG25,
  author       = {Sourjyadip Ray and
                  Shubham Sharma and
                  Somak Aditya and
                  Pawan Goyal},
  editor       = {Christos Christodoulopoulos and
                  Tanmoy Chakraborty and
                  Carolyn Rose and
                  Violet Peng},
  title        = {EduVidQA: Generating and Evaluating Long-form Answers to Student Questions
                  based on Lecture Videos},
  booktitle    = {Proceedings of the 2025 Conference on Empirical Methods in Natural
                  Language Processing, {EMNLP} 2025, Suzhou, China, November 4-9, 2025},
  pages        = {34701--34727},
  publisher    = {Association for Computational Linguistics},
  year         = {2025},
  url          = {https://doi.org/10.18653/v1/2025.emnlp-main.1760},
  doi          = {10.18653/V1/2025.EMNLP-MAIN.1760},
  bibsource    = {dblp computer science bibliography, https://dblp.org}
}

@inproceedings{gonzalez-etal-2023-automatically,
    title = "Automatically Generated Summaries of Video Lectures May Enhance Students' Learning Experience",
    author = "Gonzalez, Hannah  and
      Li, Jiening  and
      Jin, Helen  and
      Ren, Jiaxuan  and
      Zhang, Hongyu  and
      Akinyele, Ayotomiwa  and
      Wang, Adrian  and
      Miltsakaki, Eleni  and
      Baker, Ryan  and
      Callison-Burch, Chris",
    editor = {Kochmar, Ekaterina  and
      Burstein, Jill  and
      Horbach, Andrea  and
      Laarmann-Quante, Ronja  and
      Madnani, Nitin  and
      Tack, Ana{\"i}s  and
      Yaneva, Victoria  and
      Yuan, Zheng  and
      Zesch, Torsten},
    booktitle = "Proceedings of the 18th Workshop on Innovative Use of NLP for Building Educational Applications (BEA 2023)",
    month = jul,
    year = "2023",
    address = "Toronto, Canada",
    publisher = "Association for Computational Linguistics",
    url = "https://aclanthology.org/2023.bea-1.31/",
    doi = "10.18653/v1/2023.bea-1.31",
    pages = "382--393"
}

@article{DBLP:journals/tvcg/SchwabSTFHHSKKP17,
  author       = {Michail Schwab and
                  Hendrik Strobelt and
                  James Tompkin and
                  Colin Fredericks and
                  Connor Huff and
                  Dana Higgins and
                  Anton Strezhnev and
                  Mayya Komisarchik and
                  Gary King and
                  Hanspeter Pfister},
  title        = {booc.io: An Education System with Hierarchical Concept Maps and Dynamic
                  Non-linear Learning Plans},
  journal      = {{IEEE} Trans. Vis. Comput. Graph.},
  volume       = {23},
  number       = {1},
  pages        = {571--580},
  year         = {2017},
  url          = {https://doi.org/10.1109/TVCG.2016.2598518},
  doi          = {10.1109/TVCG.2016.2598518},
  bibsource    = {dblp computer science bibliography, https://dblp.org}
}

@inproceedings{DBLP:conf/chi/LiuKW18,
  author       = {Ching (Jean) Liu and
                  Juho Kim and
                  Hao{-}Chuan Wang},
  editor       = {Regan L. Mandryk and
                  Mark Hancock and
                  Mark Perry and
                  Anna L. Cox},
  title        = {ConceptScape: Collaborative Concept Mapping for Video Learning},
  booktitle    = {Proceedings of the 2018 {CHI} Conference on Human Factors in Computing
                  Systems, {CHI} 2018, Montreal, QC, Canada, April 21-26, 2018},
  pages        = {387},
  publisher    = {{ACM}},
  year         = {2018},
  url          = {https://doi.org/10.1145/3173574.3173961},
  doi          = {10.1145/3173574.3173961},
  bibsource    = {dblp computer science bibliography, https://dblp.org}
}

@article{DBLP:journals/tvcg/ZhouYCWWWCW25,
  author       = {Zhiguang Zhou and
                  Li Ye and
                  Lihong Cai and
                  Lei Wang and
                  Yigang Wang and
                  Yongheng Wang and
                  Wei Chen and
                  Yong Wang},
  title        = {ConceptThread: Visualizing Threaded Concepts in {MOOC} Videos},
  journal      = {{IEEE} Trans. Vis. Comput. Graph.},
  volume       = {31},
  number       = {2},
  pages        = {1354--1370},
  year         = {2025},
  url          = {https://doi.org/10.1109/TVCG.2024.3361001},
  doi          = {10.1109/TVCG.2024.3361001},
  bibsource    = {dblp computer science bibliography, https://dblp.org}
}

@article{breslow2013studying,
  title={Studying learning in the worldwide classroom research into edX's first MOOC.},
  author={Breslow, Lori and Pritchard, David E and DeBoer, Jennifer and Stump, Glenda S and Ho, Andrew D and Seaton, Daniel T},
  journal={Research \& Practice in Assessment},
  volume={8},
  pages={13--25},
  year={2013},
  publisher={ERIC}
}

@article{schacter2015enhancing,
  title={Enhancing attention and memory during video-recorded lectures.},
  author={Schacter, Daniel L and Szpunar, Karl K},
  journal={Scholarship of Teaching and Learning in Psychology},
  volume={1},
  number={1},
  pages={60},
  year={2015},
  publisher={Educational Publishing Foundation}
}

@article{gutierrez2024video,
  title={Video-based lecture engagement in a flipped classroom environment},
  author={Guti{\'e}rrez-Gonz{\'a}lez, Raquel and Zamarron, Alvaro and Royuela, Ana},
  journal={BMC Medical Education},
  volume={24},
  number={1},
  pages={1218},
  year={2024},
  publisher={Springer},
  doi = {10.1186/s12909-024-06228-x},
}

@inbook{Ally+2008+15+44,
url = {https://doi.org/10.15215/aupress/9781897425084.003},
title = {1. Foundations of Educational Theory for Online Learning},
booktitle = {The Theory and Practice of Online Learning},
author = {Mohamed Ally},
editor = {Terry Anderson},
publisher = {Athabasca University Press},
address = {Athabasca},
pages = {15--44},
doi = {doi:10.15215/aupress/9781897425084.003},
isbn = {9781897425077},
year = {2008},
lastchecked = {2025-08-30}
}

@inproceedings{liu-etal-2025-talk,
    title = "What Is That Talk About? A Video-to-Text Summarization Dataset for Scientific Presentations",
    author = "Liu, Dongqi  and
      Whitehouse, Chenxi  and
      Yu, Xi  and
      Mahon, Louis  and
      Saxena, Rohit  and
      Zhao, Zheng  and
      Qiu, Yifu  and
      Lapata, Mirella  and
      Demberg, Vera",
    editor = "Che, Wanxiang  and
      Nabende, Joyce  and
      Shutova, Ekaterina  and
      Pilehvar, Mohammad Taher",
    booktitle = "Proceedings of the 63rd Annual Meeting of the Association for Computational Linguistics (Volume 1: Long Papers)",
    month = jul,
    year = "2025",
    address = "Vienna, Austria",
    publisher = "Association for Computational Linguistics",
    url = "https://aclanthology.org/2025.acl-long.310/",
    doi = "10.18653/v1/2025.acl-long.310",
    pages = "6187--6210",
    ISBN = "979-8-89176-251-0"
}

@inproceedings{DBLP:conf/coling/WangCLCHSWZ25,
  author       = {Tianshu Wang and
                  Xiaoyang Chen and
                  Hongyu Lin and
                  Xuanang Chen and
                  Xianpei Han and
                  Le Sun and
                  Hao Wang and
                  Zhenyu Zeng},
  editor       = {Owen Rambow and
                  Leo Wanner and
                  Marianna Apidianaki and
                  Hend Al{-}Khalifa and
                  Barbara Di Eugenio and
                  Steven Schockaert},
  title        = {Match, Compare, or Select? An Investigation of Large Language Models
                  for Entity Matching},
  booktitle    = {Proceedings of the 31st International Conference on Computational
                  Linguistics, {COLING} 2025, Abu Dhabi, UAE, January 19-24, 2025},
  pages        = {96--109},
  publisher    = {Association for Computational Linguistics},
  year         = {2025},
  url          = {https://aclanthology.org/2025.coling-main.8/},
  bibsource    = {dblp computer science bibliography, https://dblp.org}
}

@inproceedings{DBLP:conf/emnlp/WuGGHLD25,
  author       = {Qian Wu and
                  Zheyao Gao and
                  Longfei Gou and
                  Yifan Hou and
                  Ann Sin Nga Lau and
                  Qi Dou},
  editor       = {Christos Christodoulopoulos and
                  Tanmoy Chakraborty and
                  Carolyn Rose and
                  Violet Peng},
  title        = {HealthCards: Exploring Text-to-Image Generation as Visual Aids for
                  Healthcare Knowledge Democratizing and Education},
  booktitle    = {Proceedings of the 2025 Conference on Empirical Methods in Natural
                  Language Processing, {EMNLP} 2025, Suzhou, China, November 4-9, 2025},
  pages        = {27536--27558},
  publisher    = {Association for Computational Linguistics},
  year         = {2025},
  url          = {https://doi.org/10.18653/v1/2025.emnlp-main.1401},
  doi          = {10.18653/V1/2025.EMNLP-MAIN.1401},
  bibsource    = {dblp computer science bibliography, https://dblp.org}
}

@Inbook{Meek2023,
author="Meek, Sarah E. M.
and Neve, Hilary
and Wearn, Andy",
editor="Nestel, Debra
and Reedy, Gabriel
and McKenna, Lisa
and Gough, Suzanne",
title="Threshold Concepts and Troublesome Knowledge",
bookTitle="Clinical Education for the Health Professions: Theory and Practice ",
year="2023",
publisher="Springer Nature Singapore",
address="Singapore",
pages="361--383",
isbn="978-981-15-3344-0",
doi="10.1007/978-981-15-3344-0_25",
url="https://doi.org/10.1007/978-981-15-3344-0_25"
}

@article{kalyuga2005prior,
  title={Prior knowledge principle in multimedia learning},
  author={Kalyuga, Slava},
  journal={The Cambridge handbook of multimedia learning},
  pages={325--337},
  year={2005},
  doi={10.1017/CBO9780511816819.022}
}

@inproceedings{DBLP:conf/nips/ZhengC00WZL0LXZ23,
  author       = {Lianmin Zheng and
                  Wei{-}Lin Chiang and
                  Ying Sheng and
                  Siyuan Zhuang and
                  Zhanghao Wu and
                  Yonghao Zhuang and
                  Zi Lin and
                  Zhuohan Li and
                  Dacheng Li and
                  Eric P. Xing and
                  Hao Zhang and
                  Joseph E. Gonzalez and
                  Ion Stoica},
  editor       = {Alice Oh and
                  Tristan Naumann and
                  Amir Globerson and
                  Kate Saenko and
                  Moritz Hardt and
                  Sergey Levine},
  title        = {Judging LLM-as-a-Judge with MT-Bench and Chatbot Arena},
  booktitle    = {Advances in Neural Information Processing Systems 36: Annual Conference
                  on Neural Information Processing Systems 2023, NeurIPS 2023, New Orleans,
                  LA, USA, December 10 - 16, 2023},
  year         = {2023},
  url          = {http://papers.nips.cc/paper\_files/paper/2023/hash/91f18a1287b398d378ef22505bf41832-Abstract-Datasets\_and\_Benchmarks.html},
  bibsource    = {dblp computer science bibliography, https://dblp.org}
}

@article{swellerCLT,
author = {Sweller, John and Ayres, Paul and Kalyuga, Slava},
year = {2011},
month = {01},
pages = {},
title = {Cognitive Load Theory},
isbn = {978-1-4419-8125-7},
doi = {10.1007/978-1-4419-8126-4}
}

@article{sweller2020cognitive,
  title={Cognitive load theory and educational technology},
  author={Sweller, John},
  journal={Educational technology research and development},
  volume={68},
  number={1},
  pages={1--16},
  year={2020},
  publisher={JSTOR},
  doi = {10.1007/s11423-019-09701-3}
}

@inproceedings{DBLP:conf/cvpr/LeeGC25,
  author       = {Min Jung Lee and
                  Dayoung Gong and
                  Minsu Cho},
  title        = {Video Summarization with Large Language Models},
  booktitle    = {{IEEE/CVF} Conference on Computer Vision and Pattern Recognition,
                  {CVPR} 2025, Nashville, TN, USA, June 11-15, 2025},
  pages        = {18981--18991},
  publisher    = {Computer Vision Foundation / {IEEE}},
  year         = {2025},
  url          = {https://openaccess.thecvf.com/content/CVPR2025/html/Lee\_Video\_Summarization\_with\_Large\_Language\_Models\_CVPR\_2025\_paper.html},
  doi          = {10.1109/CVPR52734.2025.01768},
  bibsource    = {dblp computer science bibliography, https://dblp.org}
}

@techreport{flux2pro,
  title       = {FLUX.2: Frontier Visual Intelligence},
  author      = {Black Forest Labs},
  institution = {Black Forest Labs},
  year        = {2025},
  url         = {https://bfl.ai/models/flux-2},
  note        = {Available at \url{https://bfl.ai/models/flux-2}}
}

@techreport{gemini3_1flashImage,
  title       = {Gemini 3.1 Flash Image Model Card},
  author      = {{Google DeepMind}},
  institution = {Google},
  year        = {2026},
  url         = {https://storage.googleapis.com/deepmind-media/Model-Cards/Gemini-3-1-Flash-Image-Model-Card.pdf},
  note        = {Available at \url{https://storage.googleapis.com/deepmind-media/Model-Cards/Gemini-3-1-Flash-Image-Model-Card.pdf}}
}

@article{DBLP:journals/ce/ChenW15,
  author       = {Chih{-}Ming Chen and
                  Chung{-}Hsin Wu},
  title        = {Effects of different video lecture types on sustained attention, emotion,
                  cognitive load, and learning performance},
  journal      = {Comput. Educ.},
  volume       = {80},
  pages        = {108--121},
  year         = {2015},
  url          = {https://doi.org/10.1016/j.compedu.2014.08.015},
  doi          = {10.1016/J.COMPEDU.2014.08.015},
  bibsource    = {dblp computer science bibliography, https://dblp.org}
}

@misc{zhao2026qwenimage20technicalreport,
      title={Qwen-Image-2.0 Technical Report}, 
      author={Bing Zhao and Chenfei Wu and Deqing Li and Hao Meng and Jiahao Li and Jie Zhang and Jingren Zhou and Junyang Lin and Kaiyuan Gao and Kuan Cao and Kun Yan and Liang Peng and Lihan Jiang and Niantong Li and Ningyuan Tang and Shengming Yin and Tianhe Wu and Xiao Xu and Xiaoyue Chen and Xihua Wang and Yan Shu and Yanran Zhang and Yi Wang and Yilei Chen and Ying Ba and Yixian Xu and Yujia Wu and Yuxiang Chen and Zecheng Tang and Zekai Zhang and Zhendong Wang and Zihao Liu and Zikai Zhou and An Yang and Chen Cheng and Chenxu Lv and Dayiheng Liu and Fan Zhou and Hantian Xiong and Hongzhu Shi and Hu Wei and Huihong Zhao and Ivy Liu and Jianwei Zhang and Jiawei Zhang and Kai Chen and Kang He and Levon Xue and Lin Qu and Linhan Tang and Luwen Feng and Minggang Wu and Minmin Sun and Na Ni and Rui Men and Shuai Bai and Sishou Zheng and Tao Lan and Tianqi Zhang and Tingkun Wen and Wei Wang and Weixu Qiao and Weiyi Lu and Wenmeng Zhou and Xiaodong Deng and Xiaoxiao Xu and Xinlei Fang and Xionghui Chen and Yanan Wang and Yang Fan and Yichang Zhang and Yixuan Xu and Yu Wu and Zhiyuan Ma and Zhizhi Cai},
      year={2026},
      eprint={2605.10730},
      archivePrefix={arXiv},
      primaryClass={cs.CV},
      url={https://arxiv.org/abs/2605.10730}, 
}

@techreport{google2025gemini3flash,
  title       = {Gemini 3 Flash Model Card},
  author      = {{Google DeepMind}},
  institution = {Google},
  year        = {2025},
  url         = {https://storage.googleapis.com/deepmind-media/Model-Cards/Gemini-3-Flash-Model-Card.pdf},
  note        = {Available at \url{https://storage.googleapis.com/deepmind-media/Model-Cards/Gemini-3-Flash-Model-Card.pdf}}
}

@inproceedings{DBLP:conf/acl/WangRWS25,
  author       = {Junling Wang and
                  Anna Rutkiewicz and
                  April Yi Wang and
                  Mrinmaya Sachan},
  editor       = {Wanxiang Che and
                  Joyce Nabende and
                  Ekaterina Shutova and
                  Mohammad Taher Pilehvar},
  title        = {Generating Pedagogically Meaningful Visuals for Math Word Problems:
                  {A} New Benchmark and Analysis of Text-to-Image Models},
  booktitle    = {Findings of the Association for Computational Linguistics, {ACL} 2025,
                  Vienna, Austria, July 27 - August 1, 2025},
  series       = {Findings of {ACL}},
  pages        = {11229--11257},
  publisher    = {Association for Computational Linguistics},
  year         = {2025},
  url          = {https://aclanthology.org/2025.findings-acl.586/},
  bibsource    = {dblp computer science bibliography, https://dblp.org}
}

@article{mayer1998cognitive,
  title={A cognitive theory of multimedia learning: Implications for design principles},
  author={Mayer, Richard E and Moreno, Roxana},
  journal={Journal of educational psychology},
  volume={91},
  number={2},
  pages={358--368},
  year={1998}
}

@book{MayerMultimediaLearning,
author = {Mayer, Richard E.},
title = {Multimedia Learning},
year = {2009},
isbn = {0521514126},
publisher = {Cambridge University Press},
address = {USA},
edition = {2nd}
}

@article{WONG2012449,
title = {Cognitive load theory, the transient information effect and e-learning},
journal = {Learning and Instruction},
volume = {22},
number = {6},
pages = {449-457},
year = {2012},
issn = {0959-4752},
doi = {https://doi.org/10.1016/j.learninstruc.2012.05.004},
author = {Anna Wong and Wayne Leahy and Nadine Marcus and John Sweller}
}

@article{gpt5,
  author       = {OpenAI},
  title        = {OpenAI {GPT-5} System Card},
  journal      = {CoRR},
  volume       = {abs/2601.03267},
  year         = {2026},
  url          = {https://doi.org/10.48550/arXiv.2601.03267},
  doi          = {10.48550/ARXIV.2601.03267},
  eprinttype   = {arXiv},
  eprint       = {2601.03267},
  bibsource    = {dblp computer science bibliography, https://dblp.org}
}

@techreport{anthropic2026claudesonnet46,
  title        = {System Card: Claude Sonnet 4.6},
  author       = {{Anthropic}},
  institution  = {Anthropic},
  year         = {2026},
  month        = feb,
  url          = {https://www.anthropic.com/claude-sonnet-4-6-system-card}
}

\clearpage
\appendix

\label{sec:appendix}

\section{Methodological Details of the \framework\ Framework}
\subsection{Extracting Concept Map}
\label{app:sec:concept_map_extraction}
Table~\ref{app:tab:prompt_concept_map} presents the step-by-step prompting methods for extracting concept maps from the transcript chunks and corresponding slide frames.

\begin{table*}[t]
\noindent
\small
\caption{Prompts for Extracting Concept Map.}
\label{app:tab:prompt_concept_map}
\begin{tabularx}{\textwidth}{X}
    \toprule
    \rowcolor{lightgrey} \rule{0pt}{11pt}\textbf{Step 1 - Concept Extraction} \\[1pt]
    \midrule
    Focus on the given chunk and the corresponding slides. The task is to extract knowledge concepts from the provided sentences in the chunk. Knowledge concepts should be informative and meaningful entities with clear semantic significance, no more than three words.\\
    
    \midrule
    \rowcolor{lightgrey} \rule{0pt}{14pt}\textbf{Step 2.1 - Relation Extraction} \\[1pt]
    \midrule
    Focus on concepts in CL, explore relationships between concepts based on the sentences in the given chunk, and the corresponding slides. Generate knowledge triples among these concepts in the format (head, relation, tail), where the 'head' and 'tail' are concepts in CL, and 'relation' represents the extracted relationship between them.\\
    
    \midrule
    \rowcolor{lightgrey} \rule{0pt}{11pt}\textbf{Step 2.2 - Checking Missing Triples} \\[1pt]
    \midrule
    Focus on all the heads and tails in this JSON and the previous JSON; do you miss any relationships among them that are indicated or mentioned in and across all the given chunks and the corresponding slides, such as inclusion and other logical relationships? \\

    \midrule
    \rowcolor{lightgrey} \rule{0pt}{11pt}\textbf{Step 2.3 - Checking Triples Content} \\[1pt]
    \midrule
    Check the triples based on the following two principles for constructing triples: Each triple should represent a complete proposition with a clear and direct relational description. The head and tail of the triple should be distinct knowledge concepts in CL, each concept should be an informative and meaningful entity with clear semantic significance, and no more than three words. If any triples do not meet these principles, refine them accordingly to ensure compliance based on the given chunk. \\

    \midrule
    \rowcolor{lightgrey} \rule{0pt}{11pt}\textbf{Step 3.1 - Entity Matching within Chunks} \\[1pt]
    \midrule
    Check this JSON for triples where the heads and tails may have different descriptions but refer to the same real-world entity. Identify duplicated entities that are functionally identical due to minor formatting, punctuation, or spelling variations (e.g., ``5 number summary'' and ``Five-number summary''). Only select entities that unambiguously represent the same object or label without conceptual nuance. Do not group technical synonyms (e.g., ``Q1'' and ``25th percentile'') where distinct terminology serves an educational purpose.\\

    \midrule
    \rowcolor{lightgrey} \rule{0pt}{11pt}\textbf{Step 3.2 - Entity Matching across Chunks} \\[1pt]
    \midrule
    Refer to triples across all chunks. Identify duplicated entities that are functionally identical due to minor formatting, punctuation, or spelling variations (e.g., ``5 number summary'' and ``Five-number summary''). Only select entities that unambiguously represent the same object or label without conceptual nuance. Do not group technical synonyms (e.g., ``Q1'' and ``25th percentile'') where distinct terminology serves an educational purpose.\\
    \bottomrule
    
\end{tabularx}
\end{table*}

\subsection{Important and Challenging Concepts Identification}
\label{app:sec:img_chg_idenfitication}
Figure~\ref{app:fig:prompt_important_concepts} shows the evaluation criteria used to score the importance of concepts, and Table~\ref {app:fig:prompt_challenging_concepts} shows the evaluation criteria used to identify the challenging concepts.

\begin{figure}[!h]
    \begin{tcolorbox}[
        title={\textbf{\footnotesize{Prompt for Scoring Concept Importance}}},
        colback=gray!5,
        colframe=darkgray!90
    ]
\footnotesize
\textbf{Step 1-Analyze:}
Please evaluate the importance of each concept based on the following four criteria:\\
\textbf{1. Multimedia Signaling:} Whether the concept is visually emphasized on slides (e.g., primary heading, large arrows, or highlighting).\\
\textbf{2. Lexicogrammatical Relevance Markers:} Whether verbal cues are used to emphasize a concept's high importance (e.g., verb patterns (``Must remember,''), noun phrases (e.g., ``The most important,'' ``the key thing''), and evaluative adjectives (e.g., ``crucial,'' ``fundamental'')).\\
\textbf{3. Structural Connectivity:} Whether the concept is pivotal to the network's connectivity (i.e., if this concept were removed, would the logical flow or the integration of other sub-topics be significantly compromised?).\\
\textbf{4. Temporal Allocation:} Whether the instructor invests significant narrative effort to define, elaborate, or revisit it across multiple segments or slides.\\
\noindent Evaluate and score each concept based on the four defined criteria.\\

\textbf{Step 2-Rank:} Rank the candidate concepts from highest to lowest weighted importance.
\end{tcolorbox}
\caption{Illustration of the prompt used to score the importance of the concepts in the concept map}
\label{app:fig:prompt_important_concepts}
\end{figure}

\begin{figure}[!h]
    \begin{tcolorbox}[
        title={\textbf{\footnotesize{Prompt for Identifying Challenging Concepts.}}},
        colback=gray!5,
        colframe=darkgray!90
    ]
\footnotesize
\textbf{Step 1-Analyze:}
Please evaluate whether each candidate concept is challenging for student Alex (a university student with little prior knowledge of the subject, politically neutral and objective) based on the following eight criteria:\\
\textbf{1. Counterintuitive:} Whether the concept goes against students’ current conceptual model of the topic.\\
\textbf{2. Inert:} Whether the concept can not fit easily into learners’ current understanding or may conflict with their world view or beliefs, e.g., a student may not understand why anyone would smoke when they know that smoking causes disease, until the student encounters someone who experiences little power or agency.\\
\textbf{3. Alien:} Whether the student cannot connect the concept to the world around them, so then struggles to apply it meaningfully in practice or in new contexts.\\
\textbf{4. Ritual:}  Whether the concept is primarily encoded as routine answers to common questions, making it difficult for students to grasp the underlying principles? For instance, students may be able to perform a respiratory examination exactly as taught, yet lack understanding of the underlying concepts and the rationale for each step.\\
\textbf{5. Troublesome language:} Whether the concept includes medical jargon which may be meaningless to students and render a topic incomprehensible.\\
\textbf{6. Conceptually difficult:} Whether the concept is complex in nature. For example, many medical concepts are extremely complex.\\
\textbf{7. Tacit:} Whether the concept is implicit within a discipline but may not be overtly explained during the video lecture.\\
\textbf{8. Nettlesome:} Whether the concept challenges cultural or individual beliefs and results in an intense emotional response.\\
\noindent Select the candidate concepts satisfying at least one of the eight criteria. Discard low-level details; retain only challenging concepts that are important for learning. \\
\noindent \textbf{Step 2-Rank:} Rank the selected challenging concepts from highest to lowest level of challenging based on the eight criteria.
\end{tcolorbox}
\caption{Illustration of the prompt used to identify challenging concepts from the concept map.}
\label{app:fig:prompt_challenging_concepts}
\end{figure}

\subsection{Knowledge Units Construction Algorithm}
\label{appendix:ku_subgraph}

Algorithm \ref{alg:knowledge_units} shows the detailed procedure for extracting subgraphs to construct knowledge units via label propagation.

\begin{algorithm}[!t]
\caption{\textbf{Extract Knowledge Unit Subgraphs from a Concept Map}}
\label{alg:knowledge_units}
\begin{algorithmic}[1]
\renewcommand{\COMMENT}[1]{{\color{gray}//\ #1}}
\renewcommand{\algorithmicrequire}{\textbf{Input:}}
\renewcommand{\algorithmicensure}{\textbf{Output:}}

\REQUIRE Concept Map $G = (V, E)$, Seed Node Set $C_{\text{seed}} = C_{\text{imp}} \cup C_{\text{chg}} \subseteq V$.
\ENSURE Set of Knowledge Units $K$

\STATE Execute Label Propagation on Concept Map $G$:
\STATE $P \leftarrow \text{LabelSpreading}(G, C_{\text{seed}})$, \quad where $P \in \mathbb{R}^{|V| \times |C_{\text{seed}}|}$

\FORALL{$v \in V$}
    \STATE $M_v \leftarrow \max_{s \in C_{\text{seed}}} P_{v,s}$ 
\ENDFOR

\FORALL{$s \in C_{\text{seed}}$}
    \STATE $V_s \leftarrow \{ v \in V \mid P_{v,s} \ge 0.8 \times M_v \} \cup \{s\}$
    \STATE $V_s \leftarrow \text{ConnectedComponent}(G[V_s])$ \COMMENT{Filter out isolated nodes}
    \STATE $K \leftarrow K \cup \{G[V_s]\}$
\ENDFOR

\STATE $K \leftarrow \{ S \in K \mid \forall S' \in K \setminus \{S\}, E(S) \not\subseteq E(S') \}$ \COMMENT{Deduplication}

\RETURN $K$
\end{algorithmic}
\end{algorithm}

Note that the LabelSpreading model is configured with a maximum of 1,000 iterations, a convergence tolerance of $10^{-3}$, and a clamping factor of $\alpha = 0.9$. Here, $\alpha = 0.9$ ensures that nodes depend more on the surrounding graph structure than on their original seed labels. This allows certain seed concepts to be absorbed by seeds, capturing the inherent hierarchical nature of knowledge, in which concepts can be subsumed or nested.

\subsection{Visual Summarization Generation}
\label{appendix:visual_gen}
We present prompts for generating visual summaries from knowledge units in Table~\ref{tab:prompt_visual}.

\begin{table*}[t]
\noindent
\small
\caption{Prompts used by \framework\ to generate the visual summaries in Figure~\ref{fig:framework}.}
\label{tab:prompt_visual}
\begin{tabularx}{\textwidth}{X}
    \toprule
    \rowcolor{lightgrey} \rule{0pt}{11pt}\textbf{Step 1: Visual Storyboard Generation} \\[1pt]
    \midrule
    Generate a concise storyboard for a visual summary explaining ``lubrication'' for a novice learner. Focus on illustrating the knowledge triples. Take the transcript and slide images as the factual basis.\\
    Please follow the steps below:\\
    1. GENRE \& STRATEGY: Choose the most appropriate genre (Magazine Style, Annotated Chart, Partitioned Poster, Flow Chart, or Comic Strip) to best illustrate the core concepts.\\
    2. OUTLINE: Generate a concise storyboard outline by synthesizing the provided knowledge triples into their most essential visual narrative arc. Be faithful to the source material. Use the minimum number of panels to capture the core logic without unnecessary details.\\
    3. PANEL BREAKDOWN: For each panel, provide:\\
    \quad $\bullet$ Panel ID: (e.g., Panel-1)\\
    \quad $\bullet$ Panel Triple To Be Focused: (Note which triples this panel focuses on)\\
    \quad $\bullet$ Panel Text To Be Rendered: (Provide the EXACT, concise text that should appear in the panel)\\
    \quad $\bullet$ Panel Visual Description: (Detailed description of the visual elements to be illustrated)\\

    \midrule
    \rowcolor{lightgrey} \rule{0pt}{11pt}\textbf{Step 2: Visual Summarization Generation} \\[1pt]
    \midrule
    Generate a visual educational summary for the concept 'lubrication' based on the given visual storyboard outline. Follow the outline strictly. DO NOT add any extra text or numbers outside the storyboard. Create a visually appealing design. Avoid serious or boring vibes; keep it vivid, approachable, and clear. Enrich the illustration with informative visual metaphors and expressive icons, without adding extra text. Render text in a legible format using a friendly font. Don't restrict to a rigid grid layout. Arrange the panels' layout flexibly to optimize narrative flow and visual balance.\\

    \midrule
    \rowcolor{lightgrey} \rule{0pt}{11pt}\textbf{Step 3: Post-Verification} \\[1pt]
    \midrule
    Analyze the provided visual summary for the following errors:\\
    1. Repetitions: Duplicated text, redundant panels, or ghosting elements.\\
    2. Logical inconsistency across elements (count, equations, cause-and-effect).\\
    If errors are found: Output ONLY a revised prompt using the Location-Based Structuring method that fixes these issues by removing duplications and correcting logical inconsistencies. If no errors: Output ONLY the word 'PASS'.\\
    \bottomrule
    
\end{tabularx}
\end{table*}

\section{Experiment Details}

\subsection{LLM-as-a-judge}
\label{appendix:llm_judge_prompt}
Figure~\ref{app:fig:llm_judge_prompt} presents the evaluation prompt and scoring rubrics we used for LLM-as-a-judge to assess the visual summaries.

\begin{figure*}[!h]
    \centering
    \caption{Prompts for LLM-as-a-judge of the generated visual summaries}
    \label{app:fig:llm_judge_prompt}
    
    \setlength{\parindent}{0pt}
    \renewcommand{\arraystretch}{1.1} 
    \resizebox{\linewidth}{!}{
        \begin{tabularx}{\textwidth}{|X|X|}
        \hline
        \textbf{Accuracy}\newline
        How accurately the visual illustration content factually align with the source video transcript and slides.\newline
        \textbf{1:} Very inaccurate. Contains major factual contradictions that fundamentally conflict the source.\newline
        \textbf{2:} Inaccurate. Contains multiple factual errors that are likely to cause significant misunderstanding.\newline
        \textbf{3:} Generally accurate. Largely correct, but contains at least one noticeable factual error.\newline
        \textbf{4:} Accurate. Highly correct, with only negligible errors in unimportant details.\newline
        \textbf{5:} Perfectly accurate. Entirely correct; every visual and textual detail is precise without error.
        &
        \textbf{Clarity}\newline
        \textbf{Definition:} How easily students can interpret the visual illustration without confusion or ambiguity.\newline
        \textbf{1:} Very unclear. Interpretation is nearly impossible.\newline
        \textbf{2:} Unclear. Significant content ambiguity or poor visual organization makes interpretation difficult.\newline
        \textbf{3:} Generally Clear. Content is generally understandable, though minor ambiguities or visual clutter may slow comprehension.\newline
        \textbf{4:} Clear. Content is easy to understand, with only negligible ambiguities that do not impact comprehension.\newline
        \textbf{5:} Very Clear. Content is entirely intuitive. Optimal visuals and zero ambiguity ensure immediate and accurate comprehension.
        \\
        \hline
        \textbf{Information Density}\newline
        The amount of information presented in the visual illustration.\newline
        \textbf{1:} Contains no important information.\newline
        \textbf{2:} Contains minimal information and lacks the most important source details.\newline
        \textbf{3:} Contains sufficient information and captures the essential takeaways.\newline
        \textbf{4:} Contains highly detailed information.\newline
        \textbf{5:} Contains overwhelmingly dense and excessively detailed information.
        &
        \textbf{Mental Effort}\newline
        How much mental effort is required for a learner to process the visual illustration.
        \textbf{1:} Fails to trigger active cognitive processing.\newline
        \textbf{2:} Minimal mental effort; the content is effortless to process.\newline
        \textbf{3:} Moderate mental effort; the content demands a manageable amount of cognitive load to process.\newline
        \textbf{4:} High mental effort; the content demands a high level of cognitive load to process.\newline
        \textbf{5:} Mental overload; the content demands excessive cognitive load, leading to mental exhaustion.
        \\
        \hline
        \end{tabularx}
    }
    \vspace{0.5cm}
\end{figure*}

\subsection{Baseline Methods}
\label{appendix:baseline_prompt}
We present the prompts used for the baseline methods to generate visual summaries in Table~\ref{app:tab:prompt_baseline}.

\begin{table*}[t]
\noindent
\small
\caption{Baseline method prompt examples for generating the visual summaries in Figure~\ref{fig:framework}.}
\label{app:tab:prompt_baseline}
\begin{tabularx}{\textwidth}{X}
    \toprule
    \rowcolor{lightgrey} \rule{0pt}{11pt}\textbf{Prompts for Baseline V2I} \\[1pt]
    \midrule
    Based on the provided video, create a visual educational illustration for the concept 'Lubrication.'\\

    \midrule
    \rowcolor{lightgrey} \rule{0pt}{11pt}\textbf{Prompts for Baseline T2I} \\[1pt]
    \midrule
    Based on the provided transcript, create a visual educational illustration for the concept 'Lubrication.'\\

    \midrule
    \rowcolor{lightgrey} \rule{0pt}{11pt}\textbf{Prompts for Baseline TS2I} \\[1pt]
    \midrule
     Based on the provided transcript and video slides, create a visual educational illustration for the concept 'Lubrication.'\\
    
    \midrule
    \rowcolor{lightgrey} \rule{0pt}{11pt}\textbf{Prompts for Baseline TS2I (COT) } \\[1pt]
    \midrule
    Step one: analyze the provided transcript and video slides to summarize the core knowledge of the concept 'Lubrication.' Step two: Based on the analysis, create a visual educational illustration for the concept 'Lubrication.'\\
    \bottomrule
    
\end{tabularx}
\end{table*}

\section{More Experimental Results and
Analysis}

\subsection{Concepts Evaluation}
\label{app:concept_eval}

\begin{table}[ht]
\centering
\small
\caption{Quantitative evaluation of our concept extraction method. \framework{} outperforms the baseline by extracting clearer and more accurate concepts.}
\renewcommand{\arraystretch}{1.15}
\begin{tabularx}{\linewidth}{Xcc}
\hline
Method & \textbf{Accuracy} $\uparrow$ & \textbf{Clarity} $\uparrow$ \\
\hline
Direct LLM Prompting & 4.072 & 3.472 \\
\textbf{\framework{}} & \textbf{4.144} & \textbf{3.888} \\
\hline
\end{tabularx}
\label{tab:supp-result-concepts}
\end{table}

We use GPT-5.4~\cite{gpt5} to compare the concept list extracted by \framework{} and direct prompting for all videos in our dataset. 
As Table~\ref{tab:supp-result-concepts} shows, our module extracts clearer, more accurate concepts than the baselines, which demonstrates that using \framework{}-extracted key concepts as a controlled variable actually benefits the baselines by providing high-quality anchors. This confirms that our controlled evaluation design is reasonable for isolating the quality of visual summary generation from differences in concept selection.

\subsection{Supplementary Analysis of Automatic Evaluation Results}
\label{app:subset_baseline}

Table~\ref{tab:baseline_results} reports the main comparison by averaging scores over all generated visual summaries for each method. Since some baseline models encountered failures or refusals (Gemini-V2I, Qwen-T2I, and Flux-T2I), we stratified the results in Table~\ref{tab:baseline_results} into different subsets based on the intersection of concepts and recalculated the average scores for each subset. As shown in Table~\ref{tab:concept_intersection_results}, our method consistently outperforms the baseline models across all baseline-specific valid subsets.

\begin{table*}[!ht]
\centering
\small
\caption{Supplementary subset comparison of automatic evaluation results across visual summary generation methods that encounter failures or refusals for some input concepts. Each subset is defined by the intersection of concepts for which visual summaries are successfully generated, and scores are averaged within each subset.}
\label{tab:concept_intersection_results}
\setlength{\tabcolsep}{4pt}
\renewcommand{\arraystretch}{1.12}

\vspace{0.5em}
\textbf{\framework{} vs. Gemini-V2I}
\vspace{0.25em}

\begin{tabularx}{0.85\textwidth}{p{0.3\textwidth}cccc}
\toprule
Method & Accuracy $\uparrow$ & Clarity $\uparrow$ & Information Density & Mental Effort \\
\midrule
Gemini-V2I & 3.564 & 4.102 & 3.335 & 2.981 \\
\framework{} & \textbf{3.732} & \textbf{4.477} & 2.852 & 2.404 \\
\bottomrule
\end{tabularx}

\vspace{0.8em}

\vspace{0.5em}
\textbf{\framework{} vs. Qwen-T2I}
\vspace{0.25em}

\begin{tabularx}{0.85\textwidth}{p{0.3\textwidth}cccc}
\toprule
Method & Accuracy $\uparrow$ & Clarity $\uparrow$ & Information Density & Mental Effort \\
\midrule
Qwen-T2I & 3.618 & 4.152 & 2.994 & 2.764 \\
\framework{} & \textbf{3.720} & \textbf{4.462} & 2.813 & 2.372 \\
\bottomrule
\end{tabularx}

\vspace{0.8em}

\vspace{0.5em}
\textbf{\framework{} vs. Flux-T2I}
\vspace{0.25em}

\begin{tabularx}{0.85\textwidth}{p{0.3\textwidth}cccc}
\toprule
Method & Accuracy $\uparrow$ & Clarity $\uparrow$ & Information Density & Mental Effort \\
\midrule
Flux-T2I & 3.078 & 3.443 & 3.047 & 3.055 \\
\framework{} & \textbf{3.732} & \textbf{4.480} & 2.846 & 2.403 \\
\bottomrule
\end{tabularx}

\end{table*}

\subsection{Supplementary Analysis of Ablation Study Results}
\label{app:ablation_video_level}

Section~\ref{sec:ablation} reports the ablation results by averaging scores over all generated visual summaries in each setting. 
Since different ablation settings yield varying numbers of visual summaries, we supplement the results in Table~\ref{tab:ablation_results} with a video-level comparison by averaging the scores within each video and then aggregating them across the dataset.
The results in Table~\ref{tab:video_level_ablation} are consistent with Table~\ref{tab:ablation_results}: each component is necessary to balance high accuracy and clarity while maintaining
optimal cognitive load.

\begin{table*}[!ht]
\centering
\small
\caption{Supplementary Analysis of Ablation Study Results at the video level comparison. Metric scores are averaged within each video to evaluate the individual contribution of each framework component to the generated visual summaries.}
\label{tab:video_level_ablation}
\vspace{0.5em}
\begin{tabularx}{0.85\textwidth}{>{\raggedright\arraybackslash}X c *{3}{S[table-format=1.3]}} 
\toprule
\midrule
{\makecell[l]{\textbf{Setting}}} & {\makecell[c]{Accuracy $\uparrow$ }} & {\makecell[c]{Clarity $\uparrow$}} & {\makecell[c]{Information\\Density}} & {\makecell[c]{Mental\\Effort}} \\
\midrule
\framework\  &                       \textbf{3.738} & \textbf{4.487} & 2.872 & 2.414 \\
w/o Concept Extraction              & 3.687 & 4.289 & 2.784 & 2.449 \\
w/o Knowledge Units Construction     & 3.685 & 4.474 & 2.877 & 2.354 \\
w/o Visual Summarization Generation  & 3.671 & 4.066 & 3.460 & 3.063  \\
\midrule
\bottomrule
\end{tabularx}

\end{table*}

\subsection{Cross-Model Validation of Automatic Evaluation}
\label{app:claude_results}

We conduct a supplementary evaluation with Claude-Sonnet-4.6~\cite{anthropic2026claudesonnet46} on a random 10\% subset of our dataset, covering 111 visual summaries from 17 videos. The results evaluated by Claude-Sonnet-4.6, shown in Table~\ref{tab:claude_baseline_results}, indicate that \framework{} achieves the highest accuracy and clarity while maintaining lower mental effort, showing a trend similar to the main results in Table~\ref{tab:baseline_results}. To further examine cross-model consistency, we calculate the agreement between GPT-5.4 and Claude-Sonnet-4.6 scores across all evaluated outputs ($N = 111 \times 7$). As shown in Table~\ref{tab:judge_consistency}, the two evaluators exhibit high within-one-score agreement across all metrics, further supporting the robustness of our automatic evaluation.

\begin{table*}[!h]
\centering
\small
\caption{Quantitative evaluation of visual summary generation methods using Claude-Sonnet-4.6 on a random 10\% subset of the dataset.}
\label{tab:claude_baseline_results}
\vspace{0.5em}

\begin{tabular}{l S[table-format=1.3] S[table-format=1.3] S[table-format=1.3] S[table-format=1.3]}
\toprule
\midrule
Method & {\textbf{Accuracy $\uparrow$}} & {\textbf{Clarity $\uparrow$}} & {\textbf{Information Density}} & {\textbf{Mental Effort}} \\
\midrule
Flux-T2I           & 2.910 & 3.190 & 2.937 & 2.856 \\ 

Qwen-T2I           & 2.991 & 3.523 & 2.730 & 2.478 \\

Gemini-T2I         & 3.559 & 3.793 & 3.306 & 3.081 \\ 

Gemini-V2I         & 3.396 & 3.784 & 3.234 & 2.910 \\ 

Gemini-TS2I        & 3.532 & 3.604 & 3.441 & 3.234 \\ 

Gemini-TS2I (COT)  & 3.577 & 3.505 & 3.703 & 3.441 \\ 
\midrule
\framework{}              & {\bfseries 3.658} & {\bfseries 4.027} & 2.702 & 2.387 \\ 
\midrule
\bottomrule
\end{tabular}
\end{table*}

\begin{table}[!h]
\centering
\small
\caption{Consistency between GPT-5.4 and Claude-Sonnet-4.6 scores on the evaluated subset ($N = 111 \times 7$).}
\label{tab:judge_consistency}
\setlength{\tabcolsep}{3pt}
\renewcommand{\arraystretch}{1.15}
\begin{tabularx}{\linewidth}{Xccc}
\toprule
Metric & \makecell{Exact\\Agreement} & \makecell{Within-1\\Agreement} & \makecell{Weighted\\Cohen's $\kappa$} \\
\midrule
Accuracy & 0.537 & 0.972 & 0.415 \\
Clarity & 0.430 & 0.950 & 0.288 \\
Information Density & 0.671 & 0.960 & 0.450 \\
Mental Effort & 0.622 & 0.945 & 0.383 \\
\bottomrule
\end{tabularx}
\end{table}

\section{Computational Experiments}
We implemented our proposed framework using Gemini-3-Flash and Gemini-3.1-Flash-Image. To ensure reproducibility, the temperature for Gemini-3-Flash was set to $0$. For Gemini-3.1-Flash-Image, we set the temperature to 1.0 to promote generation creativity and stylistic diversity, and configured the thinking level to High to leverage its maximum reasoning capabilities. The only exception is the post-verification regeneration step within our framework. In this step, the thinking level is set to Minimal to ensure the newly generated visuals strictly and honestly align with the refined script.
For automatic evaluation, we compared our approach against six state-of-the-art baselines utilizing Gemini-3.1-Flash-Image, Flux-2-pro, and Qwen-Image-2.
We also conducted an ablation study using both Gemini-3-Flash and Gemini-3.1-Flash-Image.
We used the same parameters (i.e., temperature, thinking level) for our framework and baseline methods.

The total cost incurred for API queries across all development, ablation, and evaluation experiments was approximately $ \$1000 $. In practice, processing a 40-minute video via the KnowVis pipeline takes about 30 minutes and costs 3 USD for API queries. The actual latency and cost can fluctuate depending on the information density and complexity of the lecture content.

\begin{figure*}[!htbp]
  \centering
  \includegraphics[width=\textwidth]{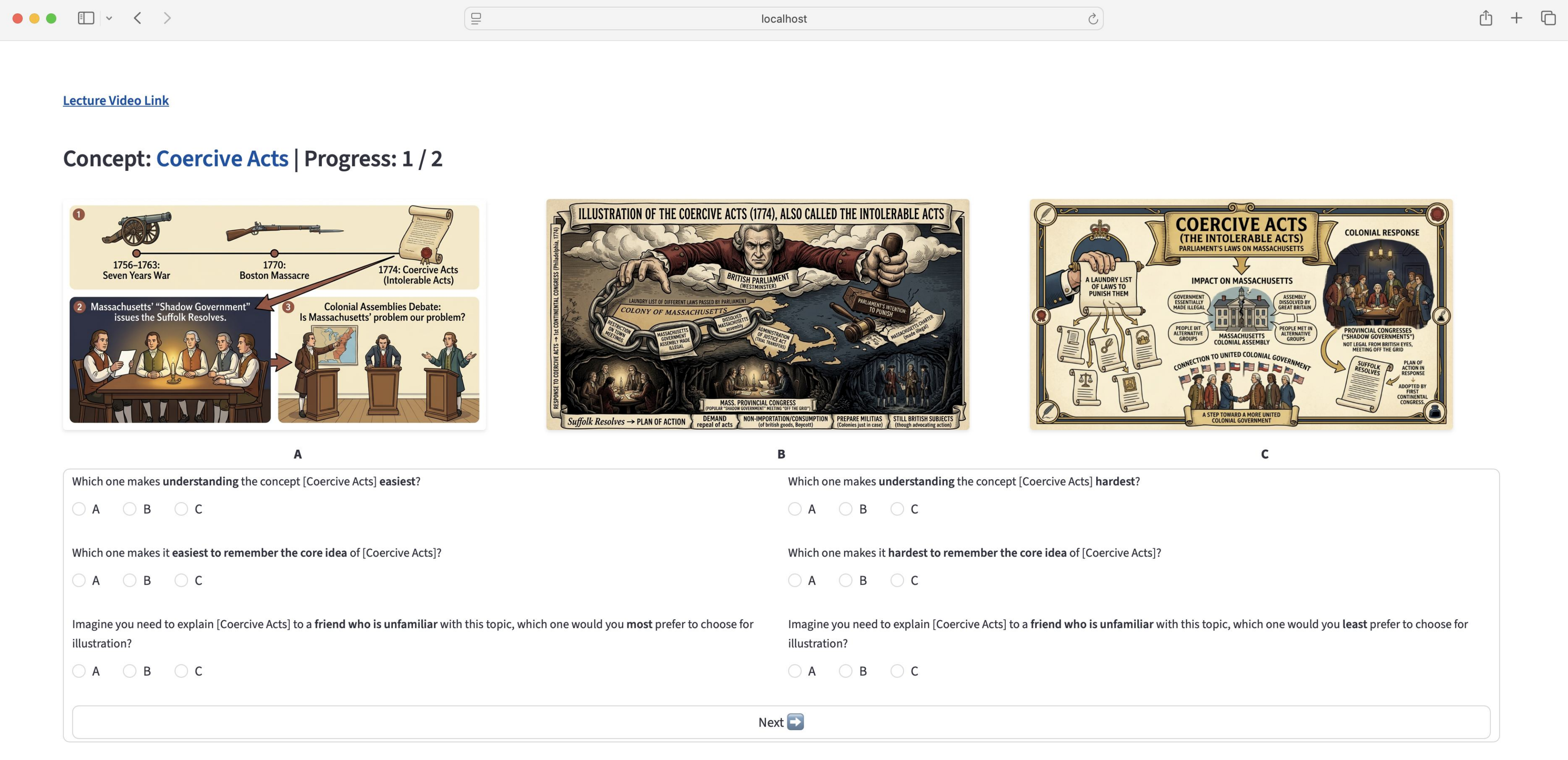}
  \hfill%
  \caption{The user interface designed for human evaluation for comparison between our generated visual summary (A) and the baselines, including Gemini-TS2V (B) and Gemini-T2V (C). To ensure an unbiased assessment, the order of the options is randomly shuffled, and participants remain blinded to the underlying methods.}
  \label{app:fig:human_eval_screenshot}
\end{figure*}
\section{Details of Human Evaluation}
\label{app:sec:human_eval}
An example of the human evaluation interface presented to the participants is provided in Figure~\ref{app:fig:human_eval_screenshot}.
Participants were recruited from a local university via word of mouth. The study was conducted remotely via Zoom, with each individual session lasting approximately 45 minutes. Each participant received a \$10 compensation after the study.

All human evaluation protocols were approved under the ethics approval ID: \texttt{HU-STA-00001449}. 
Participants were informed of the research purpose and study procedures, and they provided informed consent for the publication of their evaluation results.
No personally identifiable information was collected. Furthermore, all experimental materials were derived from open educational resources using open-access models. 
There is no risk of introducing offensive content or private data leaks.

\section{Use or Create Scientific Artifacts}
\label{app:sec:data_license_detail}

We curated our dataset by sourcing educational video lectures directly from open-access channels. These video lectures span diverse disciplines, including Physics, Astronomy, Anatomy, Psychology, Math, Sociology, Biology, Economics, U.S. History, and American Government. The curated course playlists are publicly available via the following open-access channels: 
\href{https://www.youtube.com/playlist?list=PLFqNdWpvnppJlXRzBD2UFaOt0MkjsP1oJ}{Introduction to Physical Science},
\href{https://www.youtube.com/playlist?list=PLMahwAGxKuWldeQ_qaIj8c0i7qbfyfLj4}{Introduction to Astronomy},
\href{https://www.youtube.com/@ghc-anatomyandphysiologyia6841/search?query=OpenStax\%20Anatomy\%20and\%20Physiology}{Anatomy and Physiology I and II},
\href{https://www.youtube.com/playlist?list=PLBZnw-3zC8P9LCui93rHVoynuYtSlQ0E9}{Introduction to Psychology 1e},
\href{https://oercommons.org/authoring/18167-openstax-introductory-statistics-videos-with-lectu/1/view}{Introductory Statistics},
\href{https://www.youtube.com/playlist?list=PLBZnw-3zC8P8ktS2XHus-eTY8YJWH-MtN}{Introduction to Sociology 3e}, 
\href{https://www.youtube.com/playlist?list=PLM7hZDCsTde5RI9FYRYFifkYn7kyMs_7Y}{Biology2E},
\href{https://www.youtube.com/playlist?list=PLiNGpvnwM98dDwjG7vttylclWJ2Hj8YM2}{Economics 3rd Edition},
\href{https://www.youtube.com/playlist?list=PLPvvRbQA5ODMB4vDRdRF66I9m43TMVyx3}{U.S. History Volume I: Chapters 1-16},
and \href{https://www.youtube.com/playlist?list=PLJJ7Fibh3uRPj7p-YdSNGAk4yHc3uLejE}{American Government}.
We used open educational resources and open-access models to generate visual summaries for education. 
No new content was created by the researchers.
There is no risk of introducing offensive content or personally identifying information.

The video lectures selected for our dataset are shared under permissive terms, including the Creative Commons Attribution license (CC BY, reuse allowed) and the Creative Commons Attribution — NonCommercial — ShareAlike 4.0 International License (CC BY-NC-SA 4.0), or are governed by the Standard YouTube License. Data access and content analysis were conducted strictly through YouTube's public Application Programming Interfaces (APIs) for non-commercial evaluation, ensuring full compliance with the platform's Terms of Service and academic reuse guidelines.
Our use of these scientific artifacts is consistent with their intended pedagogical and public educational purposes. 
To support future study, 
our curated dataset will be released to the research community under the CC BY-NC-SA 4.0 license.

\section{AI Assistants In Research Or Writing}
We used AI assistants such as Gemini exclusively for language editing and grammar refinement.

\newpage
\end{document}